\documentclass[letterpaper,journal]{IEEEtran}
\usepackage{amsmath,amsfonts}
\usepackage{algorithmic}
\usepackage{algorithm}
\usepackage{array}
\usepackage[caption=false,font=normalsize,labelfont=sf,textfont=sf]{subfig}
\usepackage{textcomp}
\usepackage{stfloats}
\usepackage{url}
\usepackage{verbatim}
\usepackage{graphicx}
\usepackage{cite}
\usepackage{xcolor}
\usepackage{placeins}
\usepackage[
  colorlinks=true,
  linkcolor=blue,
  citecolor=blue,
  urlcolor=blue
]{hyperref}

\begin{document}

\title{Fi-Gaussian: Frequency-Aware Implicit Gaussian Splatting for Single Image Dehazing}

\author{Yuhan~Chen,
        Ying~Fang,
        Guofa~Li,
        Wenxuan~Yu,
        Yicui~Shi,
        Kunyang~Huang,
        Wenbo~Chu,
        and Keqiang~Li%
\thanks{This work was supported by the National Natural Science Foundation of China under Grant No. 52272421. (\textit{Corresponding author: Guofa Li.})}%
\thanks{Yuhan~Chen, Ying~Fang, Guofa~Li, Wenxuan~Yu, and Yicui~Shi are with the College of Mechanical and Vehicle Engineering, Chongqing University, Chongqing 400044, China (e-mail: 20240701028@stu.cqu.edu.cn; yingfang@stu.cqu.edu.cn; liguofa@cqu.edu.cn; wenxuanyu@cqu.edu.cn; yicuishi@cqu.edu.cn).}%
\thanks{Kunyang~Huang is with the Department of Electrical and Computer Engineering, Carnegie Mellon University, Moffett Field, CA 94035, USA (e-mail: kunyangh@andrew.cmu.edu).}%
\thanks{Wenbo~Chu is with the National Innovation Center of Intelligent and Connected Vehicles, Beijing 100089, China (e-mail: chuwenbo@wicv.cn).}%
\thanks{Keqiang~Li is with the School of Vehicle and Mobility, Tsinghua University, Beijing 100084, China (e-mail: likq@tsinghua.edu.cn).}}

% The paper headers
\markboth{IEEE TRANSACTIONS ON IMAGE PROCESSING}%
{Chen \MakeLowercase{\textit{et al.}}: Fi-Gaussian: Frequency-Aware Implicit Gaussian Splatting for Single Image Dehazing}

% For IEEE journals/transactions submissions, the copyright line is usually
% added during production. Keep the following commented unless explicitly needed.
% \IEEEpubid{0000--0000/00\$00.00~\copyright~2026 IEEE}
% \IEEEpubidadjcol

\maketitle

% Placement note:
% This content does NOT belong to any \section or \subsection.
% Put it immediately after \maketitle in your IEEEtran skeleton.
% Do not add \section titles for Abstract or Index Terms.
\begin{abstract}
Single image dehazing continues to be hindered by the loss of high-frequency details and the difficulty of accurate physical scattering modeling. To address these issues, we propose Fi-Gaussian, a frequency-aware implicit Gaussian splatting network for single image dehazing. Unlike explicit rendering methods that rely on 3D point clouds, our method employs implicit Gaussian splatting to adaptively model the underlying distribution of clear images as a continuous representation in 2D feature space. The core of the network is a frequency-aware implicit Gaussian splatting module, which decouples low-frequency structural information and high-frequency texture information in the frequency domain and then performs adaptive Gaussian aggregation with complex-valued weights to recover fine details. In addition, a physics-driven scattering renormalization mechanism is introduced to estimate the transmission map and atmospheric light under the guidance of implicit Gaussian priors. Extensive experiments on multiple benchmark datasets demonstrate that Fi-Gaussian achieves state-of-the-art quantitative performance and produces visually superior dehazed results, validating the effectiveness of implicit Gaussian splatting for low-level vision tasks.
\end{abstract}


\section{Introduction}
% Placement note:
% Put the following content directly under \section{Introduction}
% in your LaTeX skeleton.
% Do not repeat \section{Introduction} here.

\IEEEPARstart{I}{n} real-world scenes, adverse weather conditions such as fog and haze cause strong light absorption and scattering in the atmosphere. This physical degradation significantly reduces image contrast and visibility, causes severe color shifts and loss of high-frequency details, and imposes substantial challenges on downstream high-level vision tasks, including autonomous driving, video surveillance, and remote sensing \cite{ayoub2025dehazingreview,agrawal2022dehazingreview,chi2023trinitynet}.
% [Citations to be filled later: 1, 3, 51]

Therefore, single image dehazing, as a fundamental inverse problem in low-level computer vision, has long attracted widespread attention from both academia and industry. Traditional dehazing methods primarily rely on natural image statistics and handcrafted priors, such as the dark channel prior (DCP)  \cite{he2010darkchannel} and the color attenuation prior (CAP) \cite{zhu2015colorattenuation}. Although these priors can produce satisfactory results in specific scenarios, their generalization ability is often limited by the highly non-uniform distribution of fog and haze in real-world scenes and the complexity of scene depth. Consequently, handcrafted priors often produce color distortion and artifacts in sky regions or under dense haze conditions \cite{ayoub2025dehazingreview,li2018reside}.
% [Citations to be filled later: 12, 49, 1, 50]

With the rapid development of deep learning, dehazing methods based on convolutional neural networks (CNNs) and Transformers have become mainstream and have achieved significant progress on large-scale benchmark datasets \cite{li2018reside,ancuti2020nhhaze}. Early end-to-end networks attempted to directly regress the transmission map or unified physical parameters in the atmospheric scattering model \cite{cai2016dehazenet,li2017aodnet,ren2016mscnn}. Subsequent studies incorporated more sophisticated modules, including multi-scale feature extraction, self-attention mechanisms, and deformable convolutions \cite{li2025tffdnet,qin2020ffanet,liu2019griddehazenet,zhu2019deformablev2,zhang2019rnan}, to improve the modeling of spatial context. To address the scarcity of paired data in real-world scenes, numerous studies have explored adversarial learning (GANs) \cite{qu2019pix2pixdehazing,deng2020hardgan,zhu2017cyclegan,engin2018cycledehaze}, synthetic-to-real domain adaptation \cite{chen2021psd,shao2020domainadaptation}, unsupervised or weakly supervised paradigms  \cite{li2021yoly,zhao2021refinednet,li2022usidnet,yang2022densitydepth}, contrastive learning, and, more recently, diffusion models and Schr\"odinger bridge theory \cite{lan2025diffusionprior,liu2025frequencydiffusion,lan2025dehazesb} to improve robustness under real-world haze distributions and facilitate physical feature decoupling  \cite{yang2018physicsdisentanglement}.
% [Citations to be filled later: 50, 52]
% [Citations to be filled later: 8, 9, 10]
% [Citations to be filled later: 2, 5, 11, 54, 55]
% [Citations to be filled later: 4, 6, 13, 14, 7, 53, 16, 17, 18, 19, 20, 21, 22, 23, 24, 15]

Nevertheless, existing deep learning-based dehazing methods still face two critical challenges. First, the strong attenuation of high-frequency components by haze causes direct spatial-domain mapping to produce overly smooth outputs with unrealistic or missing fine textures. Second, although many networks incorporate the atmospheric scattering model, the continuous distribution of the underlying physical parameters is not modeled accurately. Consequently, the transmission map and global atmospheric light cannot be adaptively disentangled during feature renormalization, which limits both the fidelity and the perceptual quality of the restored images.

Recently, 3D Gaussian splatting (3DGS) has emerged as a powerful explicit radiance field representation and has significantly advanced 3D reconstruction and novel view synthesis. Starting from the original point-based 3DGS \cite{kerbl2023gaussiansplatting} and followed by the geometrically more consistent 2DGS \cite{huang2024twodgs}, Gaussian splatting has been rapidly extended to dynamic scene modeling , large-scale urban scene rendering \cite{cheng2024gaussianpro,liu2024citygaussian,fan2025momentumgs}, and 2D-to-3D generation \cite{hanson2025speedysplat,yi2024gaussiandreamer,ni2025recondreamer}, because of its efficient continuous distribution modeling capability. More importantly, the potential of Gaussian splatting is no longer limited to 3D space. Recent Some studies have shown that Gaussian splatting on 2D planes enables high-frame-rate image compression and representation \cite{zhu2025lig,zeng2025instantgaussianimage,zhang2024gaussianimage,jiang2025beyondpixels}, supports visual-language alignment \cite{omri2025visionlanguage}, and achieves excellent image fidelity in low-level vision tasks such as arbitrary-scale image super-resolution \cite{hu2025gaussiansr} and zero-shot low-light enhancement \cite{chen2026llgaussianimage,chen2026llgaussianmap}. For restoration in adverse scattering media, recent works have incorporated 3DGS into multi-view dehazing  \cite{ma2025dehazegs,yu2025dehazegs,xu2025dehazesplat} and smoke removal reconstruction \cite{jain2025smokeseer}. However, these methods still rely on multi-view inputs or explicit point cloud initialization and therefore are not well suited to single image dehazing, where reliable depth and geometric cues are unavailable. As a result, the potential of Gaussian splatting for continuous distribution modeling in single image dehazing remains largely unexplored.
% [Citations to be filled later: 33, 34, 35, 37, 38, 39, 40, 41, 42, 36, 43, 44]
% [Citations to be filled later: 27, 28, 29, 30, 31, 32, 25, 26]
% [Citations to be filled later: 45, 46, 47, 48]

To overcome the dual limitations of high-frequency detail loss and inaccurate physical modeling, we propose Fi-Gaussian, a novel frequency-aware implicit Gaussian splatting network for single image dehazing. Unlike conventional explicit rendering methods that rely on 3D point clouds, Gaussian splatting is reformulated as an implicit continuous process in 2D feature space, enabling adaptive modeling of the underlying distribution of clear images.

Specifically, the proposed network consists of two closely coupled modules. First, the Frequency-Aware Implicit Gaussian Splatting (Fi-GS) module introduces the fast Fourier transform (FFT) to decouple low-frequency structures and high-frequency textures, considering the non-uniform suppression of different frequency components by haze. Within this module, complex-valued weights are used for dynamic implicit Gaussian aggregation across frequency bands, and an adaptive gating mechanism is introduced for spatial-frequency fusion, thereby enabling high-fidelity detail reconstruction. Second, the Physically Driven Scattering Renormalization (SRB) module is designed to address the black-box limitation of purely data-driven approaches. Under the guidance of implicit Gaussian priors, a spatially aware channel attention mechanism is introduced to estimate the transmission map and atmospheric light at multiple scales within the physical degradation model. The dehazing process is then reformulated as a physically guided feature inversion process through residual normalization.

By combining these two core mechanisms, Fi-Gaussian preserves strict physical consistency while overcoming the representation limits of discrete pixel domains. In summary, the main contributions of this work are as follows:
\begin{itemize}
    \item Gaussian splatting is introduced into single image dehazing for the first time through Fi-Gaussian, a 2D implicit Gaussian representation network that does not rely on explicit 3D point clouds, opening a new direction for low-level image restoration through continuous feature distribution modeling.
    \item A frequency-aware implicit Gaussian splatting module (Fi-GS) and a physically driven scattering renormalization module (SRB) are designed. The former addresses high-frequency detail loss through frequency-domain decoupling and dynamic aggregation with complex-valued weights, while the latter enables accurate estimation of the transmission map and atmospheric light by combining implicit Gaussian priors with attention mechanisms.
    \item Extensive experiments are conducted on multiple widely used benchmark datasets. Both quantitative and qualitative results show that Fi-Gaussian achieves state-of-the-art performance in objective metrics and visual quality, while also preserving color fidelity in haze-degraded scenes.
\end{itemize}

% [Introduction text to be filled.]
% [No figure/table/equation code inserted at this stage.]

% Placement note:
% Put the following block in your LaTeX skeleton starting at
% \section{Related Work}
% and replace the current Related Work placeholder block, including
% the two subsection placeholders under it.
%
% In other words, this block should appear after \section{Introduction}
% and before \section{Proposed Method}.

\section{Related Work}
This section first reviews the development of single image dehazing and then discusses the evolution of Gaussian splatting and its recent applications in vision tasks.

\subsection{Single Image Dehazing}
As a highly ill-posed problem, single image dehazing has attracted extensive attention over the past decade, leading to the development of a wide range of methods \cite{ayoub2025dehazingreview,agrawal2022dehazingreview}.
% [Citations to be filled later: 1, 3]

Early classical approaches primarily relied on physical degradation models, such as the atmospheric scattering model, together with handcrafted priors. For instance, He \textit{et al.} proposed the dark channel prior (DCP), which estimates the transmission map based on the statistical observation that natural clear images often contain low-intensity pixels in non-sky regions \cite{he2010darkchannel}. Zhu \textit{et al.} introduced the color attenuation prior (CAP), which constructs a scene depth model based on differences in hue and saturation \cite{zhu2015colorattenuation}. These prior-based methods established an important physical foundation for dehazing, were widely evaluated on datasets such as RESIDE \cite{li2018reside}, and provided useful insights for complex scenarios, including remote sensing images \cite{chi2023trinitynet}.
% [Citations to be filled later: 12, 49, 50, 51]

With the advent of deep learning, handcrafted feature engineering was gradually replaced by end-to-end learning frameworks. DehazeNet \cite{cai2016dehazenet}, AOD-Net \cite{li2017aodnet}, and MSCNN \cite{ren2016mscnn} were among the first methods to employ convolutional neural networks (CNNs) to regress transmission maps or unified physical parameters. To address blurred details and limited receptive fields, more sophisticated architectures were subsequently introduced. GridDehazeNet \cite{liu2019griddehazenet} adopted a multi-scale grid structure, FFA-Net \cite{qin2020ffanet} proposed a feature fusion attention mechanism, and TFFD-Net \cite{li2025tffdnet} explored a two-stage feature restoration strategy. In addition, deformable convolutions \cite{zhu2019deformablev2} and non-local attention \cite{zhang2019rnan} were widely adopted to improve the model's adaptive spatial perception under varying haze densities.
% [Citations to be filled later: 8, 9, 10, 11, 5, 2, 54, 55]

However, strongly supervised models rely heavily on paired synthetic haze-clear datasets, which often leads to severe domain shifts under real-world non-uniform haze conditions, such as those in the NH-HAZE benchmark \cite{ancuti2020nhhaze}. To address this limitation, numerous studies have shifted toward unsupervised, weakly supervised, or unpaired translation paradigms. Unpaired image translation and contrastive learning frameworks, represented by CycleGAN \cite{zhu2017cyclegan} and CUT \cite{park2020cut}, were soon introduced into dehazing, leading to methods such as Cycle-Dehaze \cite{engin2018cycledehaze}, physics-feature-decoupling approaches \cite{yang2018physicsdisentanglement}, and enhanced GAN-based models \cite{qu2019pix2pixdehazing,deng2020hardgan}. To reduce the domain gap between synthetic and real images, PSD \cite{chen2021psd} and related studies \cite{shao2020domainadaptation} incorporated domain adaptation strategies. Methods such as YOLY \cite{li2021yoly}, RefineDNet \cite{zhao2021refinednet}, and USID-Net \cite{li2022usidnet} explored decoupled feature representations and network fine-tuning without paired supervision. Furthermore, UCL-Dehaze \cite{wang2024ucldehaze} further advanced unsupervised contrastive learning for real-world dehazing, while recent methods have introduced diffusion models, density decomposition, and Schr\"odinger bridge theory into dehazing \cite{yang2022densitydepth,lan2025diffusionprior,liu2025frequencydiffusion,lan2025dehazesb}, substantially improving the generalization of the restored results.
% [Citations to be filled later: 52, 13, 20, 14, 15, 4, 6, 7, 53, 16, 17, 18, 21, 19, 22, 23, 24]

Unlike existing methods that mainly operate in the spatial pixel domain or latent space, Fi-Gaussian introduces continuous Gaussian distribution fitting at the feature level and combines frequency-domain modeling with physical priors to address these limitations.

\subsection{Gaussian Splatting and Its Applications}
Gaussian splatting was first introduced by Kerbl \textit{et al.} \cite{kerbl2023gaussiansplatting}, where 3D scenes are represented as a set of anisotropic 3D Gaussians and rendered through efficient differentiable rasterization, achieving real-time frame rates while maintaining high rendering quality. Subsequently, Huang \textit{et al.} introduced 2DGS \cite{huang2024twodgs}, a dimensionality-reduced extension that uses flattened 2D Gaussian patches to conform more accurately to geometric surfaces, thereby mitigating the multi-view inconsistencies of 3D Gaussians on complex structures. Owing to the explicit representation, flexibility, and ease of optimization of Gaussian splatting, this framework has been rapidly extended to the rendering of dynamic and large-scale urban scenes. For instance, Street Gaussians \cite{yan2024streetgaussians}, DrivingGaussian \cite{zhou2024drivinggaussian}, and ReconDreamer \cite{ni2025recondreamer} focused on spatiotemporal dynamic scenarios such as autonomous driving; PMGS \cite{xu2025pmgs} and PEGS \cite{xu2025pegs} incorporated physical and event-based information to enhance spatiotemporal modeling; CityGaussian \cite{liu2024citygaussian}, GaussianPro \cite{cheng2024gaussianpro}, Momentum-GS \cite{fan2025momentumgs}, and Speedy-Splat \cite{hanson2025speedysplat} employed progressive propagation, knowledge distillation, or sparsification strategies to improve the efficiency of large-scale scene reconstruction. In addition, GaussianDreamer \cite{yi2024gaussiandreamer} explored the bridging of 2D priors and 3DGS generation.
% [Citations to be filled later: 33，34, 35, 39, 44, 37, 38, 41, 40, 42, 36, 43]

The strong representation capability of Gaussian splatting has led to its increasing use in 2D image representation and low-level vision tasks. In image representation, GaussianImage \cite{zhang2024gaussianimage}, Instant GaussianImage \cite{zeng2025instantgaussianimage}, and LIG \cite{zhu2025lig} demonstrated that 2DGS can serve as a highly compressed and high-fidelity image representation format. This representation has also been extended to dataset distillation \cite{jiang2025beyondpixels} and cross-modal vision-language alignment \cite{omri2025visionlanguage}. In image restoration, GaussianSR \cite{hu2025gaussiansr} used Gaussian representations for arbitrary-scale super-resolution, while LL-GaussianImage \cite{chen2026llgaussianimage} and LL-GaussianMap \cite{chen2026llgaussianmap} applied 2D Gaussian splatting to low-light enhancement and demonstrated strong texture reconstruction capability.
% [Citations to be filled later: 29, 28, 27, 30, 31, 32, 25, 26]

Recently, preliminary attempts have been made to apply Gaussian splatting to scattering scenes involving haze or smoke. DehazeGS \cite{ma2025dehazegs,yu2025dehazegs} proposed a 3DGS-based dehazing algorithm for multi-image settings, DehazeSplat \cite{xu2025dehazesplat} attempted to unify unsupervised dehazing with 3DGS reconstruction, and SmokeSeer \cite{jain2025smokeseer} employed 3DGS for smoke removal. However, all these pioneering studies rely on multi-view geometric constraints, camera poses, or explicit point cloud initialization, which makes them unsuitable for single-image dehazing without depth priors. In contrast, Fi-Gaussian requires no explicit 3D point cloud support and performs frequency-aware dynamic Gaussian aggregation in the 2D feature domain through an implicit representation, which extends the application of Gaussian splatting to low-level dehazing tasks.
% [Citations to be filled later: 45, 46, 47, 48]

% Fig. 1 placeholder

% Placement note:
% Put the following block in your LaTeX skeleton starting at
% \section{Proposed Method}
% and replace the current Proposed Method placeholder block, including
% the subsection placeholders under it.
%
% In other words, this block should appear after \section{Related Work}
% and before \section{Experiments}.

\section{Proposed Method}
Single image dehazing is a highly ill-posed inverse problem in low-level vision. Traditional deep learning methods typically learn an end-to-end mapping from hazy images to clear images directly in the spatial domain. This paradigm has inherent limitations in handling severe degradation in dense haze regions and in preserving high-frequency texture details. To overcome this bottleneck, we propose Fi-Gaussian, which introduces the implicit Gaussian splatting paradigm into single image dehazing. This architecture eliminates the dependence on explicit 3D point clouds and introduces the continuous representation capability of Gaussian splatting into 2D feature space, where it is closely coupled with the physical priors of the atmospheric scattering model.

At the architectural level, Fi-Gaussian adopts an asymmetric encoder-decoder framework. Given a hazy input image, the network first maps it into a high-dimensional continuous feature space through an initial convolutional module, as defined in Eq.~\eqref{eq:initial-conv}:
% [Equation (1) to be inserted here]
% F_0 = \delta(W_{\mathrm{init}} * I_{\mathrm{hazy}} + B_{\mathrm{init}})
\begin{equation}
\label{eq:initial-conv}
\mathcal{F}_0 = \delta\!\left(W_{\mathrm{init}} * \mathcal{I}_{\mathrm{hazy}} + \mathcal{B}_{\mathrm{init}}\right)
\end{equation}
where $I_{\mathrm{hazy}} \in \mathbb{R}^{3 \times H \times W}$ denotes the input image, $W_{\mathrm{init}}$ denotes the parameters of the initial convolution kernel, $\delta(\cdot)$ denotes the GELU activation function, and $F_0 \in \mathbb{R}^{C \times H \times W}$ denotes the extracted shallow features.

To capture the global contextual information of haze distribution across multiple scales and extract hierarchical high-dimensional semantic features, the encoder consists of four downsampling stages. At each scale, conventional residual blocks are replaced by stacked Physically Driven Scattering Renormalization (SRB) modules, which progressively estimate physical scattering parameters and perform feature dehazing at different resolutions. The feature flow within the encoder can be abstracted as in Eq.~\eqref{eq:encoder-flow}:
% [Equation (2) to be inserted here]
% F_{i}^{\mathrm{enc}} = M_{\mathrm{SRB}}^{(i,\mathrm{refine})}(M_{\mathrm{SRB}}^{(i)}(D_i(F_{i-1}^{\mathrm{enc}})))
\begin{equation}
\label{eq:encoder-flow}
\mathcal{F}_{i}^{\mathrm{enc}}
=
\mathcal{M}_{\mathrm{SRB}}^{(i,\mathrm{refine})}
\!\left(
\mathcal{M}_{\mathrm{SRB}}^{(i)}
\!\left(
D_i\!\left(\mathcal{F}_{i-1}^{\mathrm{enc}}\right)
\right)
\right)
\end{equation}
where $D_i(\cdot)$ denotes a downsampling operator composed of a stride-2 convolution and batch normalization, which reduces spatial resolution while doubling the channel dimension to enlarge the network's receptive field.

At the bottleneck, which is the deepest layer of the network, the feature maps have the lowest spatial resolution and the richest global semantic information. To achieve thorough dehazing and feature reconstruction in this core region, the Fi-GS and SRB modules are stacked in an alternating manner, as shown in Eq.~\eqref{eq:bottleneck-stack}:
% [Equation (3) to be inserted here]
% F_{\mathrm{bottle}} = M_{\mathrm{Fi-GS}}(M_{\mathrm{SRB}}(M_{\mathrm{Fi-GS}}(F_{4}^{\mathrm{enc}})))
\begin{equation}
\label{eq:bottleneck-stack}
\mathcal{F}_{\mathrm{bottle}}
=
\mathcal{M}_{\mathrm{Fi\mbox{-}GS}}
\!\left(
\mathcal{M}_{\mathrm{SRB}}
\!\left(
\mathcal{M}_{\mathrm{Fi\mbox{-}GS}}
\!\left(
\mathcal{F}_{4}^{\mathrm{enc}}
\right)
\right)
\right)
\end{equation}

During decoding, the network progressively restores the spatial resolution of the features through upsampling operations. To address feature-space misalignment and semantic conflicts that often arise in conventional skip connections after channel concatenation, a dense feature fusion mechanism is introduced. This mechanism integrates multi-scale encoder features and further adaptively recalibrates the fused representations through channel attention and local feature refinement, as formulated in Eq.~\eqref{eq:decoder-flow}:
% [Equation (4) to be inserted here]
% F_{i}^{\mathrm{dec}} = M_{\mathrm{SRB}}^{(i)}(M_{\mathrm{DSF}}^{(i)}(U_i(F_{i+1}^{\mathrm{dec}}) \oplus F_{i-1}^{\mathrm{enc}}))
\begin{equation}
\label{eq:decoder-flow}
\begin{aligned}
\mathcal{F}_{i}^{\mathrm{dec}}
&=
\mathcal{M}_{\mathrm{SRB}}^{(i)}
\Bigl(
\mathcal{M}_{\mathrm{DSF}}^{(i)}
\bigl(
U_i(\mathcal{F}_{i+1}^{\mathrm{dec}})
\oplus
\mathcal{F}_{i-1}^{\mathrm{enc}}
\bigr)
\Bigr)
\end{aligned}
\end{equation}
where $U_i(\cdot)$ denotes the upsampling operator, and $\oplus$ denotes concatenation along the channel dimension. Finally, the decoder output is mapped back to the image space through a multilayer perceptron reconstruction head and combined with the original input through a global residual connection, thereby ensuring the stability of the underlying color distribution, as given in Eq.~\eqref{eq:output-reconstruction}:
% [Equation (5) to be inserted here]
% I_{\mathrm{clear}} = \mathrm{Clip}(\tanh(W_{\mathrm{tail}} * F_{0}^{\mathrm{dec}} + B_{\mathrm{tail}}) + I_{\mathrm{hazy}}, 0, 1)
\begin{equation}
\label{eq:output-reconstruction}
\begin{aligned}
\mathcal{I}_{\mathrm{clear}}
&=
\operatorname{Clmap}\!\Bigl(
\tanh\!\left(
W_{\mathrm{tail}} * \mathcal{F}_{0}^{\mathrm{dec}}
+ \mathcal{B}_{\mathrm{tail}}
\right) \\
&\qquad\qquad\quad
+ \mathcal{I}_{\mathrm{hazy}},
\, 0,\, 1
\Bigr)
\end{aligned}
\end{equation}

\begin{figure*}[!t]
\centering
\includegraphics[width=\textwidth]{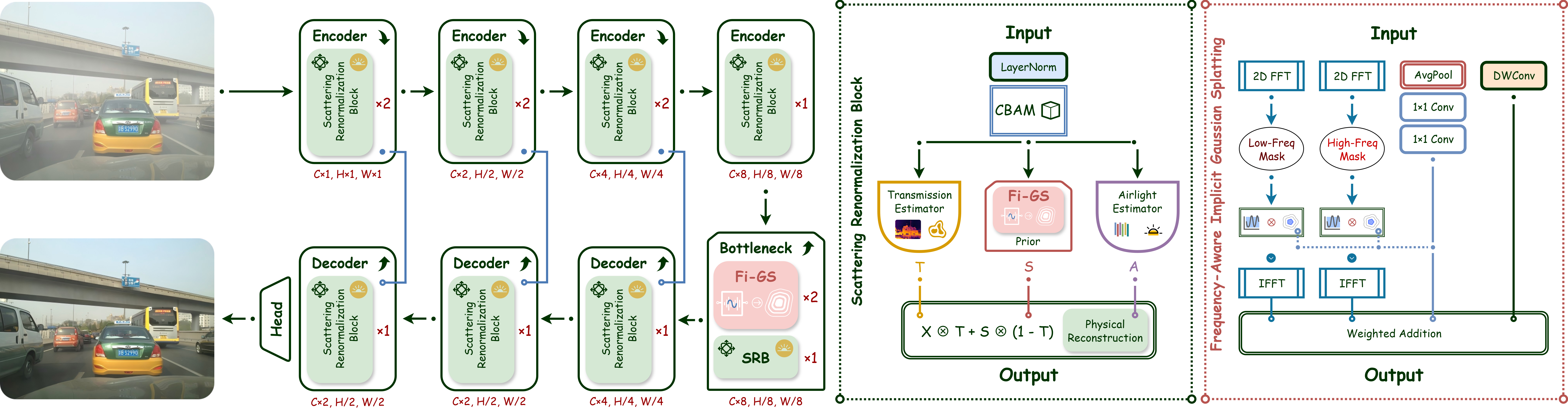}
\caption{Overall architecture of Fi-Gaussian. The model mainly consists of the Frequency-Aware Implicit Gaussian Splatting (Fi-GS) module and the Physically Driven Scattering Renormalization (SRB) module. The SRB module estimates scattering parameters under physical priors, while the Fi-GS module implicitly decouples and dynamically aggregates high- and low-frequency features in the frequency domain using complex-valued weights, thereby enabling high-fidelity detail reconstruction.}
\label{fig:fi-gaussian-architecture}
\end{figure*}

The global residual architecture enables the network to focus on learning the residual haze component, namely the haze density and distribution, rather than generating all pixels from scratch, thereby substantially improving convergence speed and generalization. The overall architecture of Fi-Gaussian is shown in Fig.~\ref{fig:fi-gaussian-architecture}.

\subsection{Rethinking Implicit Gaussian Splatting for 2D Vision}
In recent years, 3D Gaussian Splatting (3DGS) has achieved remarkable progress in novel view synthesis and scene rendering. Its core principle is to explicitly parameterize 3D space with a large set of anisotropic Gaussian ellipsoids. A standard 3D Gaussian kernel can be represented as in Eq.~\eqref{eq:3d-gaussian-kernel}:
% [Equation (6) to be inserted here]
% G_{3D}(x) = \exp\!\left(-\frac{1}{2}(x-\mu)^{T}\Sigma^{-1}(x-\mu)\right)
\begin{equation}
\label{eq:3d-gaussian-kernel}
G_{3\mathrm{D}}(x)
=
\exp\!\left(
-\frac{1}{2}
(x-\mu)^{T}\Sigma^{-1}(x-\mu)
\right)
\end{equation}

However, directly transferring explicit 3DGS to single image dehazing presents a significant theoretical gap. A single image lacks multi-view geometric constraints and therefore cannot provide a reliable 3D point cloud for initialization. Meanwhile, the explicit Gaussian rendering process incurs substantial computational overhead, which is incompatible with the high-resolution and real-time requirements of low-level vision tasks.

To address these limitations, we revisit Gaussian splatting from a theoretical perspective and propose Implicit 2D Gaussian Splatting. Gaussian splatting is reformulated from explicit coordinate-space rendering into implicit feature aggregation in the frequency domain. In a 2D continuous space, an anisotropic Gaussian kernel is defined as in Eq.~\eqref{eq:2d-gaussian-kernel}:
% [Equation (7) to be inserted here]
% g(u) = \frac{1}{2\pi |\Sigma|^{1/2}} \exp\!\left(-\frac{1}{2}(u-\mu)^{T}\Sigma^{-1}(u-\mu)\right)
\begin{equation}
\label{eq:2d-gaussian-kernel}
g(u)
=
\frac{1}{2\pi |\Sigma|^{1/2}}
\exp\!\left(
-\frac{1}{2}
(u-\mu)^{T}\Sigma^{-1}(u-\mu)
\right)
\end{equation}

According to the mathematical properties of the Fourier transform, convolution with a Gaussian function in the spatial domain is equivalent to element-wise multiplication with another Gaussian function in the frequency domain. Specifically, when Gaussian kernel convolution is applied to a feature map in the spatial domain, its frequency-domain form can be written as in Eq.~\eqref{eq:gaussian-fourier-convolution}:
% [Equation (8) to be inserted here]
% \mathcal{F}\{g(u) * X(u)\} = G(\omega) \odot \mathcal{F}\{X(u)\}
\begin{equation}
\label{eq:gaussian-fourier-convolution}
\mathcal{F}\{g(u) * X(u)\}
=
G(\omega)\odot \mathcal{F}\{X(u)\}
\end{equation}
where $G(\omega)$ denotes the Fourier transform of $g(u)$, which remains a Gaussian distribution and may include a complex phase, as shown in Eq.~\eqref{eq:gaussian-fourier-transform}:
% [Equation (9) to be inserted here]
% G(\omega) = \exp\!\left(-\frac{1}{2}\omega^{T}\Sigma \omega - j 2\pi \omega^{T}\mu\right)
\begin{equation}
\label{eq:gaussian-fourier-transform}
G(\omega)
=
\exp\!\left(
-\frac{1}{2}\omega^{T}\Sigma \omega
-
j2\pi \omega^{T}\mu
\right)
\end{equation}

Based on this mathematical equivalence, explicit computation of the covariance matrix and center coordinates of the Gaussian kernel in the spatial domain is no longer required. Instead, a set of complex-valued weight parameters $W_{\mathrm{complex}}$ can be directly learned in the frequency domain to implicitly simulate the aggregation effect of an infinite number of Gaussian kernels with adaptive variance and mean, as written in Eq.~\eqref{eq:implicit-frequency-aggregation}:
% [Equation (10) to be inserted here]
% X_{\mathrm{freq}}(\omega) = X_{\mathrm{freq}}(\omega) \odot W_{\mathrm{complex}}(\omega)
\begin{equation}
\label{eq:implicit-frequency-aggregation}
\hat{X}_{\mathrm{freq}}(\omega)
=
X_{\mathrm{freq}}(\omega)\odot W_{\mathrm{complex}}(\omega)
\end{equation}

This implicit representation offers two main advantages. First, it removes the need for explicit point cloud initialization, allowing Gaussian splatting to be integrated into fully convolutional 2D neural networks. Second, it provides full receptive-field coverage, since each point in the frequency domain contains global information from the spatial domain, which is crucial for handling large-scale non-uniform haze distributions.

\subsection{Frequency-Aware Implicit Gaussian Splatting}
Building upon the theoretical foundation of implicit Gaussian splatting, the Frequency-Aware Implicit Gaussian Splatting (Fi-GS) module is designed. The degradation patterns of hazy images exhibit strong heterogeneity in the frequency domain, with dense haze mainly manifested as smooth, large-scale variations in low-frequency components, while image edges, textures, and fine suspended particles are concentrated in high-frequency components. Conventional single-domain processing methods often produce dehazed images with blurred edges or color distortions. The Fi-GS module decouples low- and high-frequency structures in the frequency domain and applies implicit Gaussian aggregation to them separately, thereby enabling high-fidelity detail reconstruction.

Given an input feature tensor $X \in \mathbb{R}^{C \times H \times W}$, a two-dimensional real-valued fast Fourier transform (2D RFFT) is applied to map it into the complex frequency domain, as defined in Eq.~\eqref{eq:rfft-feature}:
% [Equation (11) to be inserted here]
% X_{\mathrm{freq}}(u,v) = \sum_{x=0}^{W-1} \sum_{y=0}^{H-1} X(x,y) e^{-j2\pi(ux/W + vy/H)}
\begin{equation}
\label{eq:rfft-feature}
X_{\mathrm{freq}}(u,v)
=
\sum_{x=0}^{W-1}\sum_{y=0}^{H-1}
X(x,y)\,
e^{-j2\pi\left(\frac{ux}{W}+\frac{vy}{H}\right)}
\end{equation}

Since the input is real-valued, the RFFT output $X_{\mathrm{freq}} \in \mathbb{C}^{C \times H \times (W/2+1)}$ exhibits conjugate symmetry, which greatly reduces the computational and memory costs.

To enable frequency decoupling, a dynamic mask is constructed around the center of the frequency domain, which corresponds to the low-frequency region. Let $H_{\mathrm{center}} = H/4$ and $W_{\mathrm{center}} = (W/2+1)/4$. The low-frequency mask is then defined as in Eq.~\eqref{eq:low-frequency-mask}:
% [Equation (12) to be inserted here]
% M_{\mathrm{low}}(u,v) =
% \begin{cases}
% 1, & \text{if } |u-u_c| \le W_{\mathrm{center}} \text{ and } |v-v_c| \le H_{\mathrm{center}},\\
% 0, & \text{otherwise}
% \end{cases}
\begin{equation}
\label{eq:low-frequency-mask}
\mathcal{M}_{\mathrm{low}}(u,v)
=
\begin{cases}
1, & \text{if }
\begin{aligned}[t]
&|u-u_c|\leq W_{\mathrm{center}},\\
&|v-v_c|\leq H_{\mathrm{center}},
\end{aligned}\\
0, & \text{otherwise}.
\end{cases}
\end{equation}

The high-frequency mask is defined as the complement of the low-frequency mask, as written in Eq.~\eqref{eq:high-frequency-mask}:
% [Equation (13) to be inserted here]
% M_{\mathrm{high}}(u,v) = 1 - M_{\mathrm{low}}(u,v)
\begin{equation}
\label{eq:high-frequency-mask}
\mathcal{M}_{\mathrm{high}}(u,v)
=
1-\mathcal{M}_{\mathrm{low}}(u,v)
\end{equation}

Using these masks, low- and high-frequency complex components are obtained as follows in Eq.~\eqref{eq:frequency-splitting}:
% [Equation (14) to be inserted here]
% X_{\mathrm{freq}}^{\mathrm{low}} = X_{\mathrm{freq}} \odot M_{\mathrm{low}}
% X_{\mathrm{freq}}^{\mathrm{high}} = X_{\mathrm{freq}} \odot M_{\mathrm{high}}
\begin{equation}
\label{eq:frequency-splitting}
\begin{aligned}
X_{\mathrm{freq}}^{\mathrm{low}}  &= X_{\mathrm{freq}}\odot \mathcal{M}_{\mathrm{low}},\\
X_{\mathrm{freq}}^{\mathrm{high}} &= X_{\mathrm{freq}}\odot \mathcal{M}_{\mathrm{high}}.
\end{aligned}
\end{equation}

Next, the implicit Gaussian complex weights derived in the previous subsection are introduced. In implementation, the complex weights are parameterized as tensors formed by concatenating the real and imaginary parts, resulting in learnable parameters $W_{\mathrm{low}}^{\mathrm{comp}} \in \mathbb{C}^{C \times 1 \times 1}$ and $W_{\mathrm{high}}^{\mathrm{comp}} \in \mathbb{C}^{C \times 1 \times 1}$. Implicit Gaussian aggregation is applied to the separated frequency features, as shown in Eq.~\eqref{eq:complex-weight-aggregation}:
% [Equation (15) to be inserted here]
% X_{\mathrm{freq}}^{\mathrm{low}} = X_{\mathrm{freq}}^{\mathrm{low}} \odot W_{\mathrm{low}}^{\mathrm{comp}}
% X_{\mathrm{freq}}^{\mathrm{high}} = X_{\mathrm{freq}}^{\mathrm{high}} \odot W_{\mathrm{high}}^{\mathrm{comp}}
\begin{equation}
\label{eq:complex-weight-aggregation}
\begin{aligned}
\hat{X}_{\mathrm{freq}}^{\mathrm{low}}
&=
X_{\mathrm{freq}}^{\mathrm{low}}
\odot
W_{\mathrm{low}}^{\mathrm{comp}},\\
\hat{X}_{\mathrm{freq}}^{\mathrm{high}}
&=
X_{\mathrm{freq}}^{\mathrm{high}}
\odot
W_{\mathrm{high}}^{\mathrm{comp}}.
\end{aligned}
\end{equation}

After aggregation, a two-dimensional inverse Fourier transform (2D IRFFT) maps the corrected frequency-domain representations back to the spatial domain, yielding the reconstructed low- and high-frequency features $Y_{\mathrm{low}}$ and $Y_{\mathrm{high}}$, as given in Eq.~\eqref{eq:irfft-reconstruction}:
% [Equation (16) to be inserted here]
% Y_{\mathrm{low}}(x,y) = \frac{1}{HW}\sum_{u,v} X_{\mathrm{freq}}^{\mathrm{low}}(u,v) e^{j2\pi(ux/W + vy/H)}
% Y_{\mathrm{high}}(x,y) = \frac{1}{HW}\sum_{u,v} X_{\mathrm{freq}}^{\mathrm{high}}(u,v) e^{j2\pi(ux/W + vy/H)}
\begin{equation}
\label{eq:irfft-reconstruction}
\begin{aligned}
Y_{\mathrm{low}}(x,y)
&=
\frac{1}{HW}
\sum_{u,v}
\hat{X}_{\mathrm{freq}}^{\mathrm{low}}(u,v)
e^{j2\pi\left(\frac{ux}{W}+\frac{vy}{H}\right)},\\
Y_{\mathrm{high}}(x,y)
&=
\frac{1}{HW}
\sum_{u,v}
\hat{X}_{\mathrm{freq}}^{\mathrm{high}}(u,v)
e^{j2\pi\left(\frac{ux}{W}+\frac{vy}{H}\right)}.
\end{aligned}
\end{equation}

To compensate for potential loss of local spatial continuity caused by the frequency-domain transform, a parallel convolution-based spatial enhancement branch is introduced to aggregate local information using grouped convolutions, as formulated in Eq.~\eqref{eq:spatial-branch}:
% [Equation (17) to be inserted here]
% Y_{\mathrm{spatial}} = W_{\mathrm{sp2}} * \delta(W_{\mathrm{sp1}} * X)
\begin{equation}
\label{eq:spatial-branch}
Y_{\mathrm{spatial}}
=
W_{\mathrm{sp2}} * \delta\!\left(W_{\mathrm{sp1}} * X\right)
\end{equation}

Finally, to fuse low-, high-, and spatial-frequency features, an adaptive gating-based feature aggregation mechanism is employed. The network generates normalized weight matrices for the three components through adaptive global average pooling and a multilayer perceptron, as described in Eq.~\eqref{eq:adaptive-gating}:
% [Equation (18) to be inserted here]
% \alpha_{\mathrm{gate}} = W_{\mathrm{g2}} * \delta(W_{\mathrm{g1}} * \mathrm{AvgPool}(X))
% G_{\mathrm{low}}, G_{\mathrm{high}}, G_{\mathrm{spatial}} = \mathrm{Softmax}(\alpha_{\mathrm{gate}}, \mathrm{dim}=1)
\begin{equation}
\label{eq:adaptive-gating}
\begin{aligned}
\alpha_{\mathrm{gate}}
&=
W_{\mathrm{g2}} * \delta\!\left(
W_{\mathrm{g1}} * \operatorname{AvgPool}(X)
\right),\\
\begin{bmatrix}
G_{\mathrm{low}}\\
G_{\mathrm{high}}\\
G_{\mathrm{spatial}}
\end{bmatrix}
&=
\operatorname{Softmax}\!\left(
\alpha_{\mathrm{gate}};\,\mathrm{dim}=1
\right).
\end{aligned}
\end{equation}

The final output of the Fi-GS module is obtained by adding the fused features to the input feature through a residual connection, as shown in Eq.~\eqref{eq:figs-output}:
% [Equation (19) to be inserted here]
% X_{\mathrm{out}} = X + (G_{\mathrm{low}} \odot Y_{\mathrm{low}} + G_{\mathrm{high}} \odot Y_{\mathrm{high}} + G_{\mathrm{spatial}} \odot Y_{\mathrm{spatial}})
\begin{equation}
\label{eq:figs-output}
\begin{aligned}
X_{\mathrm{out}}
=
X
&+
G_{\mathrm{low}}\odot Y_{\mathrm{low}}
+
G_{\mathrm{high}}\odot Y_{\mathrm{high}} \\
&+
G_{\mathrm{spatial}}\odot Y_{\mathrm{spatial}}.
\end{aligned}
\end{equation}

This mechanism allows the network to adaptively balance frequency-domain and spatial-domain processing across different image regions, such as smooth sky areas and complex leaf boundaries, thereby improving perceptual fidelity.

\subsection{Physics-Driven Scattering Renormalization Block}
In single image dehazing, purely data-driven black-box models are often constrained by the training-domain distribution, resulting in sharply degraded generalization under complex and variable real-world optical scattering conditions. To endow the network with explicit physical interpretability, a Physically Driven Scattering Renormalization (SRB) module is designed based on the classical atmospheric scattering model. The atmospheric scattering model in image space can be written as in Eq.~\eqref{eq:atmospheric-scattering}:
% [Equation (20) to be inserted here]
% I(x) = J(x)T(x) + A(1 - T(x))
\begin{equation}
\label{eq:atmospheric-scattering}
\mathcal{I}(x)
=
J(x)T(x) + A\bigl(1-T(x)\bigr)
\end{equation}
where $\mathcal{I}(x)$ denotes the observed hazy image, $J(x)$ denotes the latent clear image, $T(x)$ denotes the transmission map that decays exponentially with scene depth, and $A$ denotes the global atmospheric light.

The core idea of the proposed SRB is to implicitly map this pixel-domain physical model into a high-dimensional feature space, thereby leveraging the strong fitting capability of deep networks for physical parameter inversion and feature-level rendering. Given an input feature $X$, a convolutional block attention module (CBAM) is first applied to emphasize important channel and spatial features while suppressing noise, as formulated in Eq.~\eqref{eq:cbam-enhanced-feature}:
% [Equation (21) to be inserted here]
% X_{\mathrm{att}} = M_{\mathrm{spatial\_att}}(M_{\mathrm{channel\_att}}(X) \odot X)
\begin{equation}
\label{eq:cbam-enhanced-feature}
X_{\mathrm{att}}
=
\mathcal{M}_{\mathrm{spatial\_att}}
\!\left(
\mathcal{M}_{\mathrm{channel\_att}}(X)\odot X
\right)
\end{equation}

Based on this enhanced feature representation, two parallel parameter estimation branches are constructed. The first branch is used to predict multi-scale transmission maps. Considering the strong correlation between transmission and scene depth, a dilated convolution with a dilation rate of 2 is employed in this branch to enlarge the receptive field without increasing the parameter count, thereby capturing continuous global depth variations, as defined in Eq.~\eqref{eq:transmission-estimation}:
% [Equation (22) to be inserted here]
% T_{\mathrm{feat}} = \mathrm{Sigmoid}(W_{\mathrm{t3}} * \delta(W_{\mathrm{t2}}^{(\mathrm{dilated})} * \delta(W_{\mathrm{t1}} * X_{\mathrm{att}})))
\begin{equation}
\label{eq:transmission-estimation}
\resizebox{0.82\columnwidth}{!}{$
T_{\mathrm{feat}}
=
\operatorname{Sigmoid}\!\left(
W_{\mathrm{t3}} * \delta\!\left(
W_{\mathrm{t2}}^{(\mathrm{dilated})} * \delta\!\left(
W_{\mathrm{t1}} * X_{\mathrm{att}}
\right)\right)\right)
$}
\end{equation}

The second branch is used to estimate the global atmospheric light. Since atmospheric light is typically a global scalar property, adaptive global average pooling is applied to extract the global mean attribute of the feature distribution, which is then mapped to an atmospheric-light tensor through two fully connected convolutional layers, as written in Eq.~\eqref{eq:atmospheric-light-estimation}:
% [Equation (23) to be inserted here]
% A_{\mathrm{feat}} = W_{\mathrm{a2}} * \delta(W_{\mathrm{a1}} * \mathrm{AvgPool}(X_{\mathrm{att}}))
\begin{equation}
\label{eq:atmospheric-light-estimation}
A_{\mathrm{feat}}
=
W_{\mathrm{a2}} * \delta\!\left(W_{\mathrm{a1}} * \operatorname{AvgPool}(X_{\mathrm{att}})\right)
\end{equation}

After obtaining the feature-level physical scattering parameters $T_{\mathrm{feat}}$ and $A_{\mathrm{feat}}$, the crucial next step is to estimate the prior distribution of the latent clear feature $J_{\mathrm{feat}}$. Traditional physics-driven networks usually recover the latent clear feature directly from the formula, which can cause severe gradient explosion and noise amplification when [clear feature / denominator term] approaches zero under dense haze.
% [Original wording contains a missing subject before ``approaches zero''; see note in the confirmation section.]

To address this issue, the Fi-GS module is employed as an implicit clear-prior generator. The enhanced feature is fed into Fi-GS to generate a clean feature prior with reduced haze interference, as shown in Eq.~\eqref{eq:clear-prior}:
% [Equation (24) to be inserted here]
% S_{\mathrm{prior}} = M_{\mathrm{Fi-GS}}(X_{\mathrm{att}})
\begin{equation}
\label{eq:clear-prior}
S_{\mathrm{prior}}
=
\mathcal{M}_{\mathrm{Fi\mbox{-}GS}}(X_{\mathrm{att}})
\end{equation}

Subsequently, the scattering model is reconstructed in the feature space. Instead of direct mathematical division, the dehazing process is reformulated as a renormalization mechanism based on convex combination under physical parameters. The clean prior $S_{\mathrm{prior}}$ is treated as the target dehazed feature, whereas the original feature $X_{\mathrm{att}}$ retains haze-related information. According to the rewritten physical model in Eq.~\eqref{eq:rewritten-scattering}:
% [Equation (25) to be inserted here]
% J = \frac{I - A(1 - T)}{T}
\begin{equation}
\label{eq:rewritten-scattering}
J
=
\frac{\mathcal{I}-A(1-T)}{T}
\end{equation}

To maintain numerical stability and leverage residual learning, this feature recombination is equivalently implemented in the code as an interpolation weighted by the learned transmission $T_{\mathrm{feat}}$, as given in Eq.~\eqref{eq:feature-renormalization}:
% [Equation (26) to be inserted here]
% X_{\mathrm{norm}} = X_{\mathrm{att}} \odot T_{\mathrm{feat}} + S_{\mathrm{prior}} \odot (1 - T_{\mathrm{feat}})
\begin{equation}
\label{eq:feature-renormalization}
X_{\mathrm{norm}}
=
X_{\mathrm{att}}\odot T_{\mathrm{feat}}
+
S_{\mathrm{prior}}\odot (1-T_{\mathrm{feat}})
\end{equation}

\begin{figure}[!t]
\centering
\includegraphics[width=\linewidth]{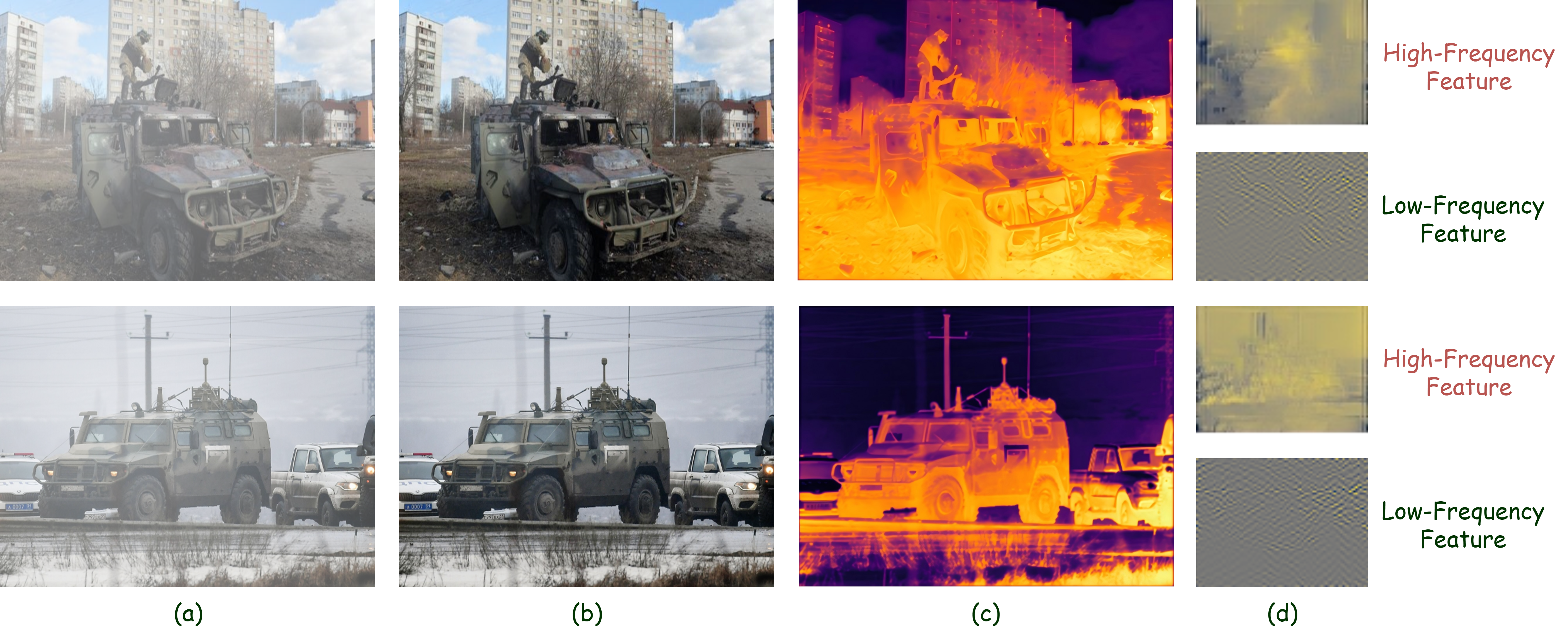}
\caption{Visualization of the internal physical and frequency-domain mechanisms of Fi-Gaussian. (a) input hazy image; (b) final dehazed result; (c) transmission map estimated by the SRB module; (d) the top and bottom images show the high-frequency textures and low-frequency structural features decoupled by the Fi-GS module in the frequency domain. This figure illustrates that implicit Gaussian aggregation preserves high-frequency details and highlights the interpretability of the physical-prior-guided representation.}
\label{fig:fi-gaussian-mechanism}
\end{figure}

This formulation has clear physical significance: when $T_{\mathrm{feat}}$ approaches 1, the network directly trusts and propagates the original input feature $X_{\mathrm{att}}$; when $T_{\mathrm{feat}}$ approaches 0, corresponding to severe haze, the network relies entirely on the clean prior $S_{\mathrm{prior}}$ generated by Fi-GS, which effectively removes low-frequency haze and restores high-frequency details. This physically driven dynamic fusion strategy effectively suppresses artifact formation. The internal physical and frequency-domain mechanisms of Fi-Gaussian are illustrated in Fig.~\ref{fig:fi-gaussian-mechanism}.

% [Figure placeholder]
% Fig. 2 caption appears in the source text after Section 3.3.
% Do not insert the figure or caption here at this stage.
% Handle Fig. 2 separately during figure backfilling.

\subsection{Multi-level Joint Constraint Loss Functions}
Achieving convergence of the implicit Gaussian splatting model in parameter space requires both a strong network architecture and comprehensive supervisory signals. Rather than relying solely on pixel-level loss, this work adopts a multi-level joint loss function that incorporates pixel-space supervision, structural awareness, frequency-domain consistency, and physical constraints.

Given the network output dehazed image $J$, the ground-truth image $I_{\mathrm{gt}}$, and the input hazy image $I_{\mathrm{hazy}}$, the total loss is defined as a weighted sum of the following components, as written in Eq.~\eqref{eq:total-loss}:
% [Equation (27) to be inserted here]
% L_{\mathrm{total}} = \lambda_1 L_{\mathrm{pix}} + \lambda_2 L_{\mathrm{ssim}} + \lambda_3 L_{\mathrm{fft}} + \lambda_4 L_{\mathrm{edge}} + \lambda_5 L_{\mathrm{grad}} + \lambda_6 L_{\mathrm{color}} + \lambda_7 L_{\mathrm{cr}}
\begin{equation}
\label{eq:total-loss}
\begin{aligned}
\mathcal{L}_{\mathrm{total}}
=
&\lambda_{1}\mathcal{L}_{\mathrm{pix}}
+\lambda_{2}\mathcal{L}_{\mathrm{ssim}}
+\lambda_{3}\mathcal{L}_{\mathrm{fft}} \\
&+\lambda_{4}\mathcal{L}_{\mathrm{edge}}
+\lambda_{5}\mathcal{L}_{\mathrm{grad}}
+\lambda_{6}\mathcal{L}_{\mathrm{color}}
+\lambda_{7}\mathcal{L}_{\mathrm{cr}}.
\end{aligned}
\end{equation}

Considering its dominant influence on PSNR, the pixel-level reconstruction loss forms the backbone of network optimization. To balance global smoothness and outlier penalization, a hybrid L1-L2 reconstruction loss $L_{\mathrm{pix}}$ is employed, as formulated in Eq.~\eqref{eq:pixel-loss}:
% [Equation (28) to be inserted here]
% L_1 = \frac{1}{N}\sum_i \lVert J_i - I_{\mathrm{gt},i} \rVert_1
% L_2 = \frac{1}{N}\sum_i \lVert J_i - I_{\mathrm{gt},i} \rVert_2^2
% L_{\mathrm{pix}} = L_1 + 0.5 \times L_2
\begin{equation}
\label{eq:pixel-loss}
\begin{aligned}
&L_{1} = \frac{1}{N}\sum_{i}\left\lVert J_{i}-I_{\mathrm{gt},i}\right\rVert_{1},\\
&L_{2} = \frac{1}{N}\sum_{i}\left\lVert J_{i}-I_{\mathrm{gt},i}\right\rVert_{2}^{2},\\
&\mathcal{L}_{\mathrm{pix}} = L_{1} + 0.5\,L_{2}.
\end{aligned}
\end{equation}

To better align the generated images with the perception of the human visual system (HVS), a structural similarity loss $L_{\mathrm{ssim}}$ based on Gaussian sliding windows is introduced. Unlike conventional local averaging, a Gaussian kernel $\omega$ is used to compute local mean $\mu$ and variance $\sigma^2$, as defined in Eq.~\eqref{eq:ssim-loss}:
% [Equation (29) to be inserted here]
% L_{\mathrm{ssim}} = 1 - \frac{1}{N}\sum \frac{(2\mu_J \mu_{\mathrm{gt}} + C_1)(2\sigma_{J,\mathrm{gt}} + C_2)}{(\mu_J^2 + \mu_{\mathrm{gt}}^2 + C_1)(\sigma_J^2 + \sigma_{\mathrm{gt}}^2 + C_2)}
\begin{equation}
\label{eq:ssim-loss}
\mathcal{L}_{\mathrm{ssim}}
=
1-
\frac{1}{N}
\sum
\frac{
(2\mu_{J}\mu_{\mathrm{gt}}+C_{1})(2\sigma_{J,\mathrm{gt}}+C_{2})
}{
(\mu_{J}^{2}+\mu_{\mathrm{gt}}^{2}+C_{1})(\sigma_{J}^{2}+\sigma_{\mathrm{gt}}^{2}+C_{2})
}
\end{equation}

Since the core design of the proposed network is frequency-aware, a frequency-domain consistency loss $L_{\mathrm{fft}}$ is introduced to ensure that implicit Gaussian aggregation does not distort the global spectral distribution. The Fourier transform of the network output is constrained by the ground-truth magnitude spectrum, as written in Eq.~\eqref{eq:fft-loss}:
% [Equation (30) to be inserted here]
% L_{\mathrm{fft}} = \lVert \mathcal{F}(J) - \mathcal{F}(I_{\mathrm{gt}}) \rVert_1
\begin{equation}
\label{eq:fft-loss}
\mathcal{L}_{\mathrm{fft}}
=
\left\lVert
\left|\mathcal{F}(J)\right|
-
\left|\mathcal{F}\!\left(\mathcal{I}_{\mathrm{gt}}\right)\right|
\right\rVert_{1}
\end{equation}

Haze significantly attenuates high-frequency image information. To preserve sharp object boundaries, a Laplacian-based edge loss $L_{\mathrm{edge}}$ is introduced. A $5 \times 5$ approximate Laplacian kernel $K_{\mathrm{lap}}$ is used to extract second-order derivative information in the spatial domain, as shown in Eq.~\eqref{eq:edge-loss}:
% [Equation (31) to be inserted here]
% L_{\mathrm{edge}} = \lVert K_{\mathrm{lap}} * J - K_{\mathrm{lap}} * I_{\mathrm{gt}} \rVert_1
\begin{equation}
\label{eq:edge-loss}
\mathcal{L}_{\mathrm{edge}}
=
\left\lVert
K_{\mathrm{lap}} * J - K_{\mathrm{lap}} * \mathcal{I}_{\mathrm{gt}}
\right\rVert_{1}
\end{equation}

To further enhance detail preservation in all directions, a gradient loss $L_{\mathrm{grad}}$ based on first-order derivatives is incorporated, as formulated in Eq.~\eqref{eq:gradient-loss}:
% [Equation (32) to be inserted here]
% \nabla_x J = J_{i,j+1} - J_{i,j}
% \nabla_y J = J_{i+1,j} - J_{i,j}
% L_{\mathrm{grad}} = \frac{1}{2}\left(\lVert \nabla_x J - \nabla_x I_{\mathrm{gt}} \rVert_1 + \lVert \nabla_y J - \nabla_y I_{\mathrm{gt}} \rVert_1\right)
\begin{equation}
\label{eq:gradient-loss}
\begin{gathered}
\nabla_{x}J = J_{i,j+1}-J_{i,j},\\
\nabla_{y}J = J_{i+1,j}-J_{i,j},\\
\mathcal{L}_{\mathrm{grad}}
=
\frac{1}{2}
\left(
\left\lVert \nabla_{x}J-\nabla_{x}\mathcal{I}_{\mathrm{gt}} \right\rVert_{1}
+
\left\lVert \nabla_{y}J-\nabla_{y}\mathcal{I}_{\mathrm{gt}} \right\rVert_{1}
\right).
\end{gathered}
\end{equation}

In dehazing tasks, contrast enhancement may induce color distortion, such as skies appearing overly blue or yellow, due to shifts in atmospheric-light color. To address this issue, a color consistency loss $L_{\mathrm{color}}$ is introduced. Global average pooling is applied to each RGB channel to compute the mean distributions, and the mean squared error is used to enforce globally consistent color tones, as defined in Eq.~\eqref{eq:color-loss}:
% [Equation (33) to be inserted here]
% L_{\mathrm{color}} = \mathrm{MSE}\!\left(\frac{1}{HW}\sum_{x,y} J(x,y), \frac{1}{HW}\sum_{x,y} I_{\mathrm{gt}}(x,y)\right)
\begin{equation}
\label{eq:color-loss}
\resizebox{0.80\columnwidth}{!}{$
\mathcal{L}_{\mathrm{color}}
=
\operatorname{MSE}\!\left(
\frac{1}{HW}\sum_{x,y} J(x,y),\,
\frac{1}{HW}\sum_{x,y} \mathcal{I}_{\mathrm{gt}}(x,y)
\right)
$}
\end{equation}

To reduce the distance between the generated and ground-truth image in a high-dimensional semantic space while pushing the hazy input away, a contrastive loss $L_{\mathrm{cr}}$ based on a pretrained VGG19 network is employed. Let $\Phi_i(\cdot)$ denote the feature extracted before the $i$-th pooling layer of VGG19 and let $\omega_i = \{1/32, 1/16, 1/8\}$ denote the corresponding weighting coefficients. This loss guides the network to reduce the distribution gap in the feature space and learn more realistic texture priors, as expressed in Eq.~\eqref{eq:contrastive-loss}:
% [Equation (34) to be inserted here]
% d_{\mathrm{ap}}^{(i)} = \lVert \Phi_i(J) - \Phi_i(I_{\mathrm{gt}}) \rVert_1
% d_{\mathrm{an}}^{(i)} = \lVert \Phi_i(J) - \Phi_i(I_{\mathrm{hazy}}) \rVert_1
% L_{\mathrm{cr}} = \sum_i \omega_i \left( \frac{d_{\mathrm{ap}}^{(i)}}{d_{\mathrm{an}}^{(i)} + \epsilon} \right)
\begin{equation}
\label{eq:contrastive-loss}
\begin{aligned}
d_{\mathrm{ap}}^{(i)}
&=
\left\lVert
\Phi_{i}(J)-\Phi_{i}(\mathcal{I}_{\mathrm{gt}})
\right\rVert_{1},\\
d_{\mathrm{an}}^{(i)}
&=
\left\lVert
\Phi_{i}(J)-\Phi_{i}(\mathcal{I}_{\mathrm{hazy}})
\right\rVert_{1},\\
\mathcal{L}_{\mathrm{cr}}
&=
\sum_{i}\omega_{i}
\left(
\frac{d_{\mathrm{ap}}^{(i)}}{d_{\mathrm{an}}^{(i)}+\epsilon}
\right).
\end{aligned}
\end{equation}

Under this multi-level loss framework, Fi-Gaussian exhibits stable optimization stability and theoretical consistency, which helps alleviate the long-standing challenges of high-frequency detail loss and inaccurate physical-parameter inversion in image dehazing. With the above derivations and implementation details, the proposed method achieves consistent improvements in both objective metrics and visual quality.

% Placement note:
% Put the following block in your LaTeX skeleton starting at
% \section{Experiments}
% and replace the current Experiments placeholder block, including
% the subsection placeholders under it.
%
% In other words, this block should appear after \section{Proposed Method}
% and before \section{Conclusion}.

\section{Experiments}

\subsection{Experimental Setup}
\indent\textbf{Dataset.} To comprehensively evaluate the performance of Fi-Gaussian in terms of objective fidelity and subjective visual quality, four representative dehazing datasets are selected, with the training and testing samples split at a strict 9:1 ratio. These datasets are divided into two groups. The first group includes the SOTS \cite{li2018reside} and NID \cite{chi2023trinitynet} datasets. SOTS is divided into indoor and outdoor scenes, while NID is categorized into light-haze and dense-haze subsets. These datasets provide paired hazy and clear reference images and are therefore suitable for full-reference evaluation of pixel-level reconstruction accuracy and structural fidelity. The second group consists of RTTS \cite{ancuti2020nhhaze} and Haze2020 \cite{shao2020domainadaptation} datasets, which mainly contain complex real-world hazy scenes without corresponding clear reference images. These datasets are therefore used for no-reference evaluation to assess the model's generalization ability and the naturalness of the generated images under real degraded conditions.
% [Citations to be filled later: 50, 51, 52, 53]

\indent\textbf{Implementation Details.} Fi-Gaussian is implemented in PyTorch and trained and evaluated on a single NVIDIA RTX 3090 GPU. The base channel dimension is set to 64. The core Fi-GS module decouples frequency components via 2D FFT and employs learnable complex weights with a standard deviation of 0.02 for dynamic feature aggregation. The model is trained for 1000 epochs with a base batch size of 8 and gradient accumulation over 4 steps. The AdamW optimizer is used with a weight decay of $10^{-4}$ and a grouped learning rate strategy: $10^{-4}$ for the encoder and $2 \times 10^{-4}$ for the decoder. After 10 epochs of linear warm-up, the learning rate is smoothly decayed using a cosine schedule to $10^{-7}$.

Training images are cropped to $256 \times 256$. In addition to standard geometric and color augmentations, Mixup is applied with a 30\% probability after the warm-up period. The total loss jointly optimizes the pixel-level reconstruction loss (L1 + L2), SSIM, FFT, and edge-gradient losses, together with other perceptual and physical objectives. Moreover, an exponential moving average (EMA) with a momentum of 0.999 is used to update the model weights during training, and test-time augmentation is applied during inference to further improve dehazing fidelity.

\indent\textbf{Evaluation Metrics.} Two evaluation protocols are employed according to the dataset type. For the SOTS and NID datasets, which contain paired reference images, seven widely used metrics are adopted, including three full-reference and four no-reference metrics. Full-reference evaluation measures the consistency between the restored images and the reference images using PSNR, SSIM, and LPIPS. PSNR is used to assess pixel-level fidelity. SSIM evaluates similarity in luminance, contrast, and structure. LPIPS measures perceptual similarity in the deep feature space of a pretrained VGG network.

To assess naturalness, contrast, and information preservation in unpaired or real-world scenarios, four no-reference metrics are employed: NIQE, LOE, DE, and EME. NIQE quantifies naturalness by measuring the statistical distance between restored and natural images; LOE evaluates the preservation of natural appearance; DE measures the richness of image information; and EME quantifies local contrast variation.

For the real-world datasets RTTS and Haze2020, which lack reference images, four advanced no-reference metrics are used: FID, NIQE, MUSIQ, and MANIQA. FID measures the distance between generated and real images in the feature space; NIQE quantifies naturalness based on natural scene statistics; MUSIQ employs a multi-scale Transformer to capture cross-scale image quality; and MANIQA uses a multi-dimensional attention mechanism to assess both local and global perceptual quality. Together, these metrics effectively reflect the perceptual quality of the model outputs in complex real-world environments.

\subsection{Performance Comparison}
We compare Fi-Gaussian with five supervised state-of-the-art (SOTA) dehazing methods \cite{qin2020ffanet,chen2021psd,cai2016dehazenet,li2017aodnet,liu2019griddehazenet} and eight unsupervised SOTA dehazing methods \cite{he2010darkchannel,engin2018cycledehaze,li2021yoly,zhao2021refinednet,li2022usidnet,yang2022densitydepth,lan2025diffusionprior,lan2025dehazesb}. The comparison results on the four selected datasets are reported in Tables~\ref{tab:nid-results}--\ref{tab:rtts-haze2020-results}.
% [Citations to be filled later: 5, 7, 8, 9, 11, 12, 14, 16, 17, 18, 19, 22, 24]

\begin{figure}[!t]
\centering
\includegraphics[width=\linewidth]{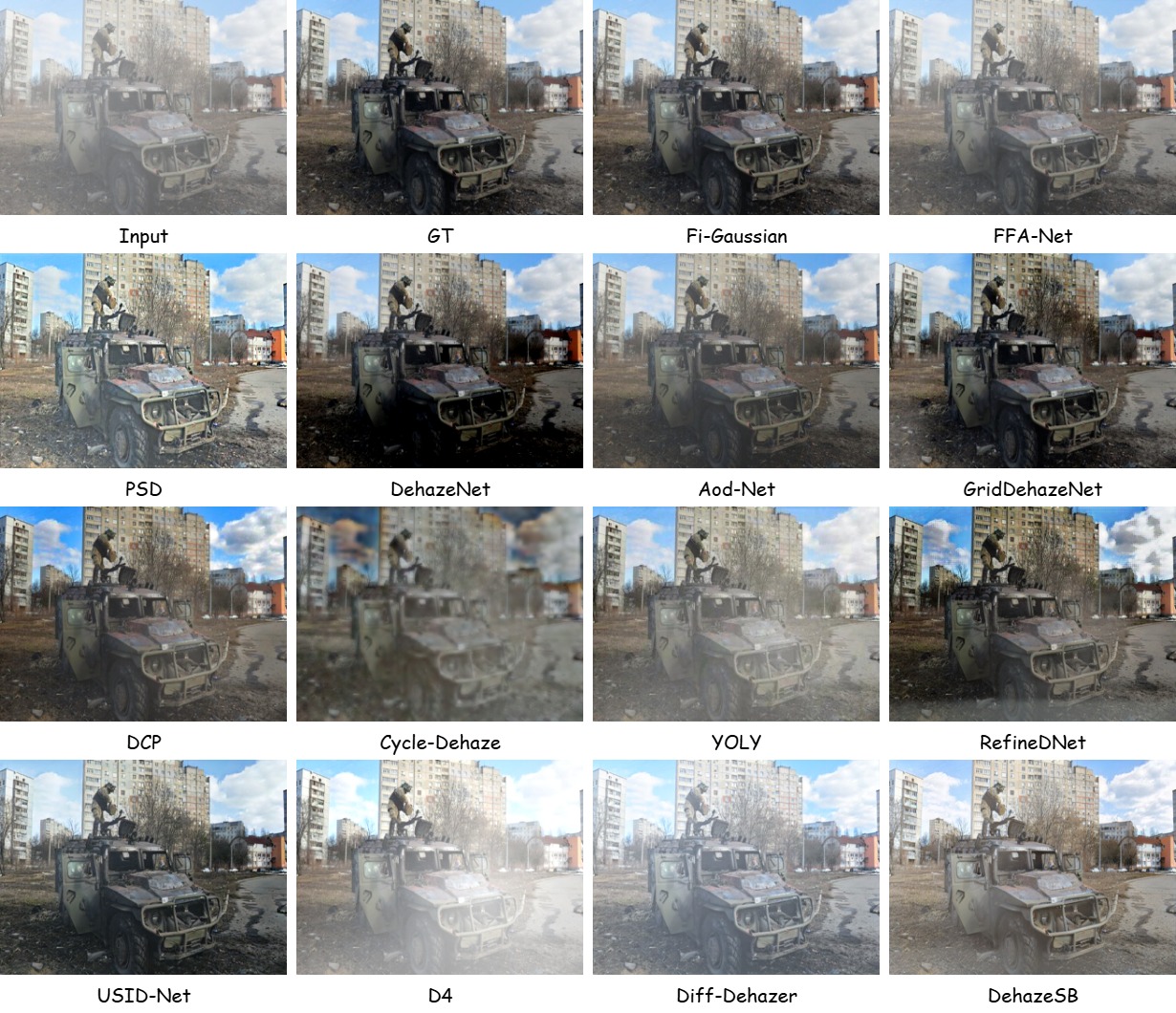}
\caption{Visual comparison of Fi-Gaussian with state-of-the-art methods on the NID dataset.}
\label{fig:nid-visual-comparison}
\end{figure}

\indent\textbf{Qualitative Comparison.} As shown in Fig.~\ref{fig:nid-visual-comparison}, visual comparisons are presented using images from the dense-haze subset of the NID dataset. In terms of overall dehazing performance, FFA-Net, AOD-Net, Cycle-Dehaze, YOLY, D4, and Diff-Dehazer exhibit clear failures, such as incomplete haze removal or uneven dehazing. In contrast, Fi-Gaussian produces visually convincing results, with clear advantages in dehazing effectiveness and overall color rendering. In terms of overall clarity, DehazeNet and GridDehazeNet produce overly high contrast, while USID-Net and Fi-Gaussian restore feature details more faithfully, yielding visually clear and well-balanced results. In terms of overall color fidelity, USID-Net, DCP, and Fi-Gaussian deliver the best visual performance.

% [Fig. 3 to be inserted later]
% [Table I to be inserted later]

\begin{figure}[!t]
\centering
\includegraphics[width=\linewidth]{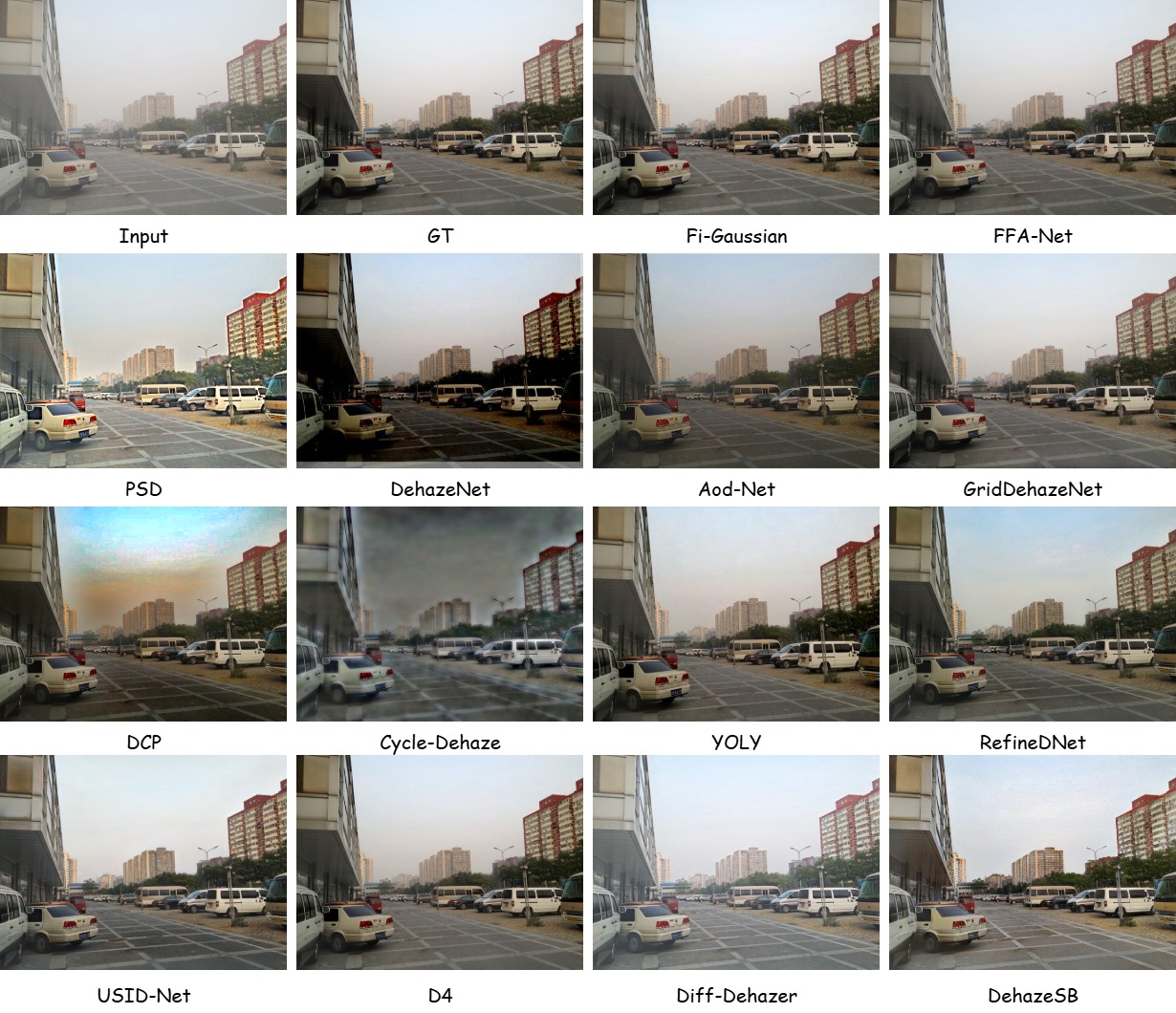}
\caption{Visual comparison of Fi-Gaussian with state-of-the-art methods on the SOTS dataset.}
\label{fig:sots-visual-comparison}
\end{figure}

Visual comparisons on images from the outdoor subset of the SOTS dataset are shown in Fig.~\ref{fig:sots-visual-comparison}. Severe color shifts are observed in the results produced by DCP and Cycle-Dehaze. In addition, DehazeNet, AOD-Net, and RefineDNet exhibit varying degrees of over-enhanced contrast and color bias. In terms of clarity, USID-Net, D4, YOLY, DehazeSB, and Fi-Gaussian all produce competitive visual results, whereas Diff-Dehazer, PSD, and FFA-Net show insufficient dehazing and leave residual haze in certain regions.

% [Fig. 4 to be inserted later]
% [Table II to be inserted later]

\begin{figure*}[!t]
\centering
\includegraphics[width=\textwidth]{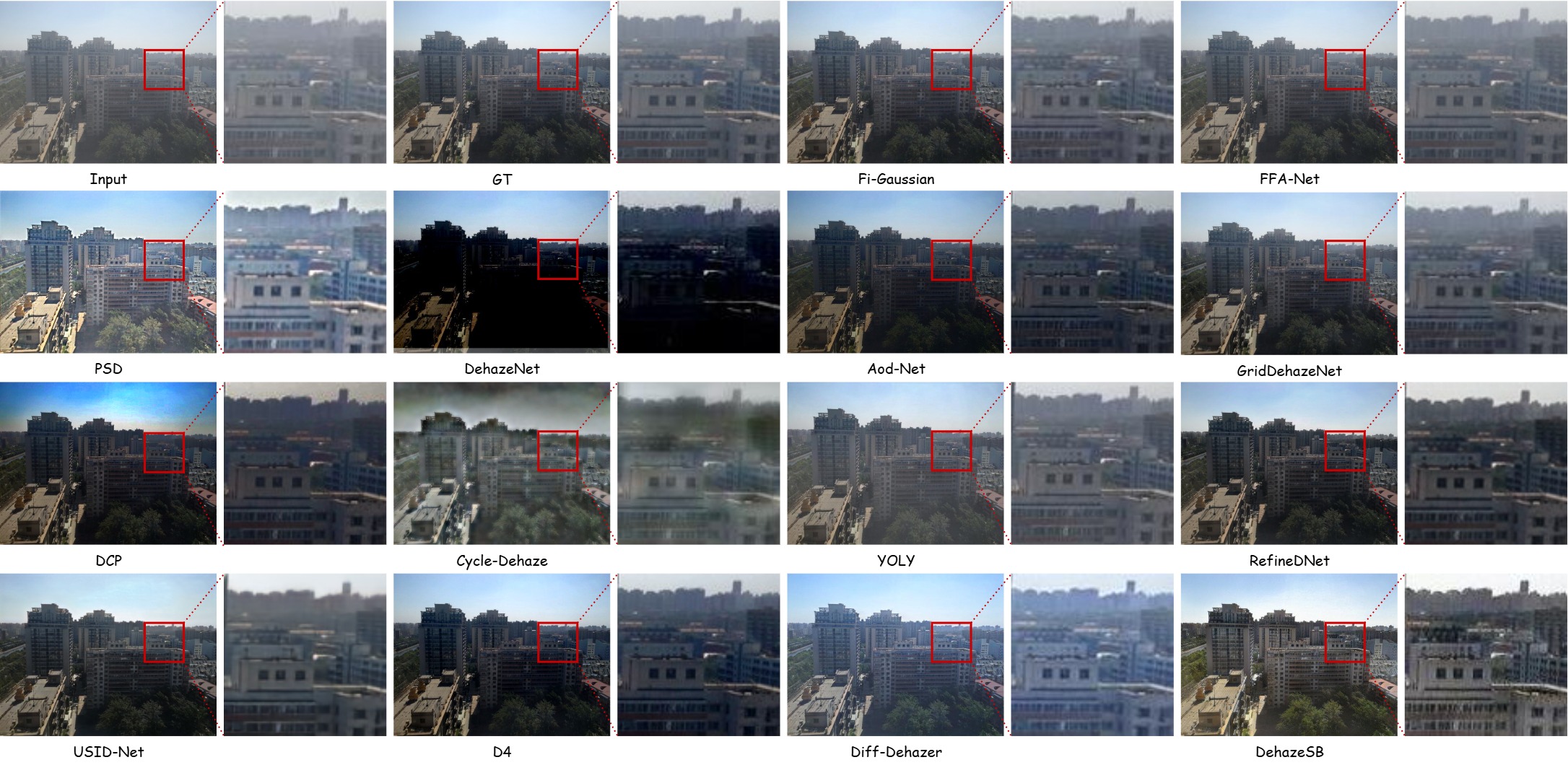}
\caption{Detail comparison of Fi-Gaussian with state-of-the-art methods on the SOTS dataset. Enlarged views of the regions marked by red boxes are provided for all methods.}
\label{fig:sots-detail-comparison}
\end{figure*}

\indent\textbf{Detail Restoration Assessment.} To further evaluate the detail-preservation capability of Fi-Gaussian, more complex scenes from the SOTS dataset are selected for fine-detail visualization. As shown in Fig.~\ref{fig:sots-detail-comparison}, Fi-Gaussian remains highly consistent with the ground truth in terms of building details, color saturation, and contrast. Other methods, however, exhibit local color deviations or overly enhanced contrast.
% [Fig. 5 to be inserted later]

Overall, in both global visual quality and fine-detail restoration, Fi-Gaussian consistently outperforms existing SOTA methods and demonstrates superior feature recovery and dehazing capability.

\indent\textbf{Quantitative Metrics Assessment.} For the paired datasets NID and SOTS, full-reference metrics (PSNR and SSIM) and no-reference metrics (NIQE, LOE, DE, and EME) are used. As shown in Table~\ref{tab:nid-results}, Fi-Gaussian ranks among the top two across all metrics on the paired NID dataset. Similarly, Table~\ref{tab:sots-results} shows that Fi-Gaussian ranks among the top two across all metrics on the SOTS dataset. For the unpaired datasets RTTS and Haze2020, Fi-Gaussian ranks among the top two in FID, NIQE, and MANIQA, as shown in Table~\ref{tab:rtts-haze2020-results}. For MUSIQ, Fi-Gaussian ranks fourth and third on the two datasets, respectively.
% [Table I to be inserted later]
% [Table II to be inserted later]
% [Table III to be inserted later]
\begin{table}[!t]
\caption{Evaluation Results of Fi-Gaussian and State-of-the-Art Methods on the NID Dataset. Red and Blue Indicate the Best and Second-Best Results for Each Metric, Respectively.}
\label{tab:nid-results}
\centering
\scriptsize
\setlength{\tabcolsep}{3pt}
\resizebox{\columnwidth}{!}{%
\begin{tabular}{lccccccc}
\hline
Method & SSIM$\uparrow$ & PSNR$\uparrow$ & LPIPS$\downarrow$ & NIQE$\downarrow$ & LOE$\downarrow$ & DE$\uparrow$ & EME$\uparrow$ \\
\hline
FFA-Net & 0.93 & \textcolor{blue}{25.82} & \textcolor{blue}{0.02} & 3.13 & 58.49 & 0.53 & 10.24 \\
PSD & 0.81 & 15.93 & 0.12 & \textcolor{red}{3.07} & 118.62 & 0.15 & 11.86 \\
DehazeNet & 0.83 & 18.39 & 0.08 & 3.33 & 89.34 & 0.51 & \textcolor{red}{20.18} \\
AOD-Net & 0.91 & 21.53 & 0.07 & 3.34 & 96.08 & 0.35 & 4.99 \\
GridDehazeNet & 0.84 & 18.17 & 0.12 & 3.13 & 110.67 & \textcolor{blue}{0.58} & 17.12 \\
DCP & 0.86 & 19.88 & 0.21 & 3.87 & 96.52 & 0.13 & 4.63 \\
Cycle-Dehaze & 0.79 & 14.82 & 0.26 & 4.28 & 189.63 & 0.08 & 3.63 \\
YOLY & 0.83 & 18.3 & 0.13 & 3.62 & 96.51 & 0.15 & 3.91 \\
RefineDNet & 0.9 & 19.36 & 0.11 & 3.59 & 84.28 & 0.23 & 7.87 \\
USID-Net & 0.89 & 22.56 & 0.12 & 3.1 & 88.45 & 0.45 & 6.53 \\
D4 & \textcolor{blue}{0.95} & 23.53 & 0.04 & 3.86 & \textcolor{blue}{49.62} & 0.39 & 5.35 \\
Diff-Dehazer & 0.76 & 11.66 & 0.2 & 4.29 & 102.68 & 0.07 & 2.55 \\
DehazeSB & 0.8 & 14.7 & 0.21 & 4.27 & 108.87 & 0.22 & 3.9 \\
Fi-Gaussian (Ours) & \textcolor{red}{0.98} & \textcolor{red}{28.56} & \textcolor{red}{0.01} & \textcolor{blue}{3.09} & \textcolor{red}{47.57} & \textcolor{red}{0.85} & \textcolor{blue}{19.0} \\
\hline
\end{tabular}%
}
\end{table}

\begin{table}[!t]
\caption{Evaluation Results of Fi-Gaussian and State-of-the-Art Methods on the SOTS Dataset. Red and Blue Indicate the Best and Second-Best Results for Each Metric, Respectively.}
\label{tab:sots-results}
\centering
\scriptsize
\setlength{\tabcolsep}{3pt}
\resizebox{\columnwidth}{!}{%
\begin{tabular}{lccccccc}
\hline
Method & SSIM$\uparrow$ & PSNR$\uparrow$ & LPIPS$\downarrow$ & NIQE$\downarrow$ & LOE$\downarrow$ & DE$\uparrow$ & EME$\uparrow$ \\
\hline
FFA-Net & 0.92 & \textcolor{blue}{32.08} & \textcolor{blue}{0.05} & 2.68 & \textcolor{blue}{21.92} & 0.43 & 3.44 \\
PSD & 0.77 & 16.02 & 0.16 & 3.23 & 141.57 & 0.06 & 4.84 \\
DehazeNet & 0.57 & 12.73 & 0.45 & 4.83 & 68.69 & 0.51 & 12.92 \\
AOD-Net & 0.86 & 17.24 & 0.13 & 3.34 & 34.32 & 0.34 & 7.8 \\
GridDehazeNet & \textcolor{blue}{0.93} & 27.17 & 0.06 & 2.7 & 31.41 & 0.49 & 5.3 \\
DCP & 0.79 & 14.71 & 0.18 & 2.63 & 55.08 & 0.5 & 8.48 \\
Cycle-Dehaze & 0.83 & 12.74 & 0.29 & 3.2 & 206.86 & 0.47 & 3.54 \\
YOLY & 0.53 & 19.08 & 0.22 & 2.98 & 135.12 & 0.11 & 2.87 \\
RefineDNet & 0.9 & 20.25 & 0.13 & 3.1 & 85.06 & 0.35 & 5.48 \\
USID-Net & 0.75 & 21.4 & 0.17 & 3.91 & 55.33 & 0.41 & 5.38 \\
D4 & 0.83 & 18.68 & 0.07 & \textcolor{red}{2.49} & 56.48 & 0.46 & \textcolor{blue}{13.63} \\
Diff-Dehazer & 0.91 & 23.38 & 0.11 & 2.69 & 41.16 & 0.51 & 4.16 \\
DehazeSB & 0.84 & 21.74 & 0.17 & 2.8 & 121.22 & \textcolor{red}{0.62} & 7.03 \\
Fi-Gaussian (Ours) & \textcolor{red}{0.98} & \textcolor{red}{33.07} & \textcolor{red}{0.02} & \textcolor{blue}{2.57} & \textcolor{red}{13.35} & \textcolor{blue}{0.52} & \textcolor{red}{14.46} \\
\hline
\end{tabular}%
}
\end{table}

\begin{table*}[!t]
\caption{Evaluation Results of Fi-Gaussian and State-of-the-Art Methods on the RTTS and Haze2020 Datasets. Red and Blue Indicate the Best and Second-Best Results for Each Metric, Respectively.}
\label{tab:rtts-haze2020-results}
\centering
\scriptsize
\renewcommand{\arraystretch}{1.12}
\setlength{\tabcolsep}{0pt}
\begin{tabular*}{\textwidth}{@{\extracolsep{\fill}}lcccccccc@{}}
\hline
Method & \multicolumn{4}{c}{RTTS} & \multicolumn{4}{c}{Haze2020} \\
\cline{2-9}
& FID$\downarrow$ & NIQE$\downarrow$ & MUSIQ$\uparrow$ & MANIQA$\uparrow$
& FID$\downarrow$ & NIQE$\downarrow$ & MUSIQ$\uparrow$ & MANIQA$\uparrow$ \\
\hline
FFA-Net & 62.0 & 3.98 & 53.2 & 0.17 & 72.9 & 3.81 & \textcolor{blue}{60.1} & 0.21 \\
PSD & 65.3 & 4.04 & 50.1 & 0.14 & 79.5 & 3.92 & 55.0 & 0.19 \\
DehazeNet & 79.3 & 4.75 & 47.2 & 0.11 & 89.4 & 4.07 & 52.9 & 0.09 \\
AOD-Net & 73.8 & 3.92 & 51.3 & 0.12 & 85.3 & 3.86 & 54.2 & 0.15 \\
GridDehazeNet & 69.3 & 3.77 & 49.7 & 0.17 & 80.2 & 3.95 & 53.7 & 0.15 \\
DCP & 74.2 & 4.31 & 50.4 & 0.12 & 81.7 & 3.81 & 53.4 & 0.12 \\
Cycle-Dehaze & 88.7 & 4.92 & 46.7 & 0.11 & 91.3 & 4.27 & 52.3 & 0.11 \\
YOLY & 71.3 & 4.55 & 52.3 & 0.13 & 84.5 & 3.88 & 54.7 & 0.13 \\
RefineDNet & 64.2 & 4.21 & 56.3 & 0.17 & 87.3 & 3.98 & 55.3 & 0.11 \\
USID-Net & 61.8 & 4.10 & 57.7 & 0.14 & 85.2 & 3.96 & 53.2 & 0.15 \\
D4 & 70.5 & 3.96 & 58.3 & 0.15 & 84.5 & 3.91 & 53.5 & 0.17 \\
Diff-Dehazer & \textcolor{red}{52.3} & 3.84 & \textcolor{blue}{59.4} & \textcolor{red}{0.33} & 70.9 & \textcolor{blue}{3.57} & \textcolor{red}{61.3} & \textcolor{red}{0.38} \\
DehazeSB & 53.1 & \textcolor{blue}{3.74} & \textcolor{red}{61.3} & 0.17 & \textcolor{blue}{69.8} & 3.64 & 59.7 & 0.18 \\
Fi-Gaussian (Ours) & \textcolor{blue}{52.8} & \textcolor{red}{3.51} & 58.7 & \textcolor{blue}{0.23} & \textcolor{red}{67.9} & \textcolor{red}{3.44} & 59.3 & \textcolor{blue}{0.27} \\
\hline
\end{tabular*}
\end{table*}

Overall, with its novel architecture and strong performance, Fi-Gaussian demonstrates both methodological novelty and competitive effectiveness and remains highly competitive against various state-of-the-art dehazing methods.

\subsection{Ablation Study}
To systematically validate the effectiveness and internal mechanisms of the core components in Fi-Gaussian, extensive ablation experiments were conducted on the NID dataset. All variant models were trained under the same strategy as the baseline model, which is a basic auto-encoder architecture retaining only standard convolutional residual blocks, to ensure a fair comparison.

\indent\textbf{Effectiveness of Core Components.} To examine the synergistic effects of the SRB and Fi-GS modules, a baseline auto-encoder architecture was constructed, with only basic convolutional residual blocks retained. Subsequently, the SRB and Fi-GS modules were added sequentially to observe the corresponding performance gains, as illustrated in Fig.~\ref{fig:fi-gaussian-architecture}.
% [Fig. 1 is referenced here and should be handled separately during figure backfilling.]

% [Table IV to be inserted later]
\begin{table}[!t]
\caption{Ablation Study of the Core Components of Fi-Gaussian. Red and Blue Indicate the Best and Second-Best Results for Each Metric, Respectively.}
\label{tab:core-components-ablation}
\centering
\footnotesize
\renewcommand{\arraystretch}{1.08}
\setlength{\tabcolsep}{0pt}
\begin{tabular*}{\columnwidth}{@{\extracolsep{\fill}}lccccc@{}}
\hline
Method & Baseline & +SRB & +Fi-GS & PSNR$\uparrow$ & SSIM$\uparrow$ \\
\hline
Baseline & $\surd$ & $\times$ & $\times$ & 24.27 & 0.84 \\
Baseline+SRB & $\surd$ & $\surd$ & $\times$ & \textcolor{blue}{27.45} & \textcolor{blue}{0.93} \\
Baseline+Fi-GS & $\surd$ & $\times$ & $\surd$ & 26.76 & 0.91 \\
Baseline+SRB+Fi-GS & $\surd$ & $\surd$ & $\surd$ & \textcolor{red}{30.48} & \textcolor{red}{0.98} \\
\hline
\end{tabular*}
\end{table}

As shown in Table~\ref{tab:core-components-ablation}, both components play critical roles and provide substantial performance gains for Fi-Gaussian. The full framework yields the largest improvement. On the NID dataset, PSNR increases by 6.21 dB and SSIM improves by 14\% relative to the baseline. When the two modules are combined, the network gains physical interpretability and recovers high-frequency texture details through implicit Gaussian splatting. This result demonstrates the strong complementarity between physics-driven modeling and implicit Gaussian rendering.

\indent\textbf{Internal Mechanisms of Fi-GS.} To analyze the contribution of the key components in Fi-GS, a systematic ablation study was conducted on frequency decoupling, complex-valued weighting, and adaptive gating fusion. Specifically, the following variants were considered: (1) removing low- and high-frequency decoupling while retaining only global FFT processing, to evaluate the necessity of explicit frequency separation; (2) replacing complex-valued weights with real-valued weights, to verify the role of complex-domain weighting in implicit Gaussian aggregation; and (3) removing the spatial branch and preserving only frequency-domain features, to assess the importance of adaptive fusion between spatial and frequency representations; and (4) removing adaptive gating and replacing it with simple feature concatenation, to assess the role of adaptive gating in spatial-frequency feature fusion.

% [Table V to be inserted later]
\begin{table}[!t]
\caption{Ablation Study of the Internal Components of the Fi-GS Module. Red and Blue Indicate the Best and Second-Best Results for Each Metric, Respectively.}
\label{tab:figs-module-ablation}
\centering
\footnotesize
\renewcommand{\arraystretch}{1.08}
\setlength{\tabcolsep}{0pt}
\begin{tabular*}{\columnwidth}{@{\extracolsep{\fill}}lcc@{}}
\hline
Variants of Fi-GS & PSNR$\uparrow$ & SSIM$\uparrow$ \\
\hline
w/o Freq-Decoupling & 29.74 & \textcolor{blue}{0.97} \\
w/o Complex Weights & 30.07 & \textcolor{red}{0.98} \\
w/o Spatial Branch & 29.08 & \textcolor{blue}{0.97} \\
w/o Adaptive Gate & \textcolor{blue}{30.11} & \textcolor{red}{0.98} \\
Fi-GS (Ours) & \textcolor{red}{30.48} & \textcolor{red}{0.98} \\
\hline
\end{tabular*}
\end{table}

As reported in Table~\ref{tab:figs-module-ablation}, removing frequency decoupling results in a 0.74 dB drop in PSNR, which indicates that separating global illumination and local textures in the frequency domain effectively prevents color shifts and excessive edge smoothing. Replacing complex-valued weights with real-valued weights disrupts phase information and thereby affects structural alignment in the spatial domain. Moreover, the adaptive fusion of spatial and frequency-domain features is critical, as it enables the network to dynamically balance spatial semantics and frequency-domain details according to the local haze density. These results validate the rationality of the Fi-GS design.

\indent\textbf{Design of SRB.} To validate the nonlinear extension of the traditional atmospheric scattering model and the feature recalibration mechanism, the internal components of the SRB module were sequentially ablated: (1) removing the physical-formula-based reconstruction of transmission and atmospheric light and replacing it with direct residual addition; (2) replacing multi-scale dilated convolution with single-scale convolution; and (3) replacing the improved spatial-channel CBAM attention with conventional SE attention.

\begin{table}[!t]
\caption{Ablation Study of the Proposed SRB Module. Red and Blue Indicate the Best and Second-Best Results for Each Metric, Respectively.}
\vspace{1pt}
\label{tab:srb-module-ablation}
\centering
\footnotesize
\renewcommand{\arraystretch}{1.08}
\setlength{\tabcolsep}{0pt}
\begin{tabular*}{\columnwidth}{@{\extracolsep{\fill}}lcc@{}}
\hline
Configuration & PSNR$\uparrow$ & SSIM$\uparrow$ \\
\hline
Direct Residual & 29.51 & 0.95 \\
Single-scale T-estimator & 29.77 & \textcolor{blue}{0.96} \\
SE Attention & \textcolor{blue}{30.07} & \textcolor{red}{0.98} \\
Ours & \textcolor{red}{30.48} & \textcolor{red}{0.98} \\
\hline
\end{tabular*}
\end{table}

As shown in Table~\ref{tab:srb-module-ablation}, the use of physical priors, multi-scale convolutions, and attention mechanisms consistently yield better results, which confirms the effectiveness of the SRB design.

\indent\textbf{Ablation on the Objective Function.} To analyze the contribution of each supervisory term in the composite loss function set to the final image quality, the loss terms were incrementally added starting from the base loss $L_{\mathrm{pix}}$. As reported in Table~\ref{tab:loss-combination-ablation}, when all loss functions are incorporated during training, Fi-Gaussian achieves a 6.13 dB improvement in PSNR compared with using only the base loss $L_{\mathrm{pix}}$, which verifies the effectiveness of the proposed multi-level loss design.
% [Table VII to be inserted later]
\begin{table}[!t]
\caption{Ablation Study of Different Loss Function Combinations. Red and Blue Indicate the Best and Second-Best Results for Each Metric, Respectively.}
\vspace{-1pt}
\label{tab:loss-combination-ablation}
\centering
\scriptsize
\setlength{\tabcolsep}{3pt}
\resizebox{\columnwidth}{!}{%
\begin{tabular}{ccccccccc}
\hline
$\mathcal{L}_{pix}$ & $\mathcal{L}_{ssim}$ & $\mathcal{L}_{fft}$ & $\mathcal{L}_{edge}$ & $\mathcal{L}_{grad}$ & $\mathcal{L}_{color}$ & $\mathcal{L}_{cr}$ & PSNR$\uparrow$ & SSIM$\uparrow$ \\
\hline
$\surd$ & $\times$ & $\times$ & $\times$ & $\times$ & $\times$ & $\times$ & 24.35 & 0.8 \\
$\surd$ & $\surd$ & $\times$ & $\times$ & $\times$ & $\times$ & $\times$ & 27.11 & 0.91 \\
$\surd$ & $\surd$ & $\surd$ & $\times$ & $\times$ & $\times$ & $\times$ & 28.32 & \textcolor{blue}{0.96} \\
$\surd$ & $\surd$ & $\surd$ & $\surd$ & $\times$ & $\times$ & $\times$ & 28.9 & \textcolor{blue}{0.96} \\
$\surd$ & $\surd$ & $\surd$ & $\surd$ & $\surd$ & $\times$ & $\times$ & 30.18 & \textcolor{blue}{0.96} \\
$\surd$ & $\surd$ & $\surd$ & $\surd$ & $\surd$ & $\surd$ & $\times$ & \textcolor{blue}{30.24} & \textcolor{red}{0.98} \\
$\surd$ & $\surd$ & $\surd$ & $\surd$ & $\surd$ & $\surd$ & $\surd$ & \textcolor{red}{30.48} & \textcolor{red}{0.98} \\
\hline
\end{tabular}%
}
\end{table}

\FloatBarrier
\section{Conclusion}
In this paper, we propose Fi-Gaussian and introduce the implicit Gaussian splatting paradigm into single image dehazing. To address the bottlenecks of high-frequency detail loss and inaccurate physical modeling in traditional methods, the proposed Frequency-Aware Implicit Gaussian Splatting (Fi-GS) module is used to enhance high-fidelity feature reconstruction. This module enables high-fidelity texture reconstruction in a continuous 2D feature space through frequency-domain decoupling and dynamic aggregation with complex-valued weights. Furthermore, the network integrates a Physically Driven Scattering Renormalization (SRB) module, which leverages implicit priors to accurately estimate the transmission map and atmospheric light. Extensive experiments demonstrate that Fi-Gaussian achieves state-of-the-art performance on multiple benchmark datasets, with clear improvements in objective metrics and visual quality. This work provides a new perspective on image dehazing and suggests the broader applicability of implicit Gaussian representations to low-level vision tasks.

\vspace*{-35pt}
% Author Imformation
\newcommand{\biogap}{\vspace{-18pt}}
\newcommand{\biogaplarge}{\vspace{-18pt}}

\begin{IEEEbiography}[{\includegraphics[width=1in,height=1.25in,clip,keepaspectratio]{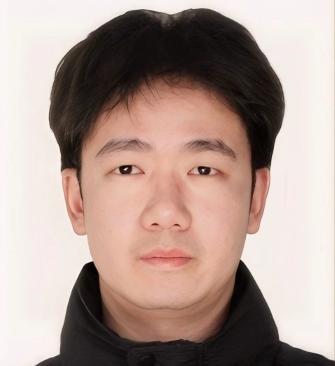}}]{Yuhan Chen}
received his master's degree in 2024 from the College of Mechanical Engineering at Chongqing University of Technology. He is currently pursuing the Ph.D. degree in College of Mechanical and Vehicle Engineering at Chongqing University, China. His research interests include deep learning, low-level vision and Gaussian splatting.
\end{IEEEbiography}
\vspace{-30pt}

\begin{IEEEbiography}[{\includegraphics[width=1in,height=1.25in,clip,keepaspectratio]{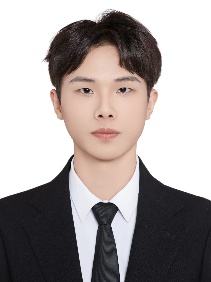}}]{Ying Fang}
received the B.E. degree in vehicle engineering from Chongqing University of Technology. He is currently pursuing the M.E. degree in mechanical engineering with Chongqing University, Chongqing, China. His research interests include computer vision, Gaussian splatting and deep learning.
\end{IEEEbiography}

\begin{IEEEbiography}[{\includegraphics[width=1in,height=1.25in,clip,keepaspectratio]{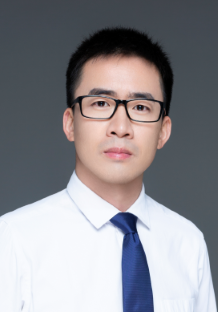}}]{Guofa Li}
received the Ph.D. degree in mechanical engineering from Tsinghua University, China, in 2016. He is currently a Professor with Chongqing University, China. His research interests include environment perception, driver behavior analysis, and smart decision-making based on artificial intelligence technologies in autonomous vehicles and intelligent transportation systems. He serves as an Associate Editor for the IEEE Transactions on Intelligent Transportation Systems, IEEE Transactions on Affective Computing, and IEEE Sensors Journal.
\end{IEEEbiography}
\biogaplarge

\begin{IEEEbiography}[{\includegraphics[width=1in,height=1.25in,clip,keepaspectratio]{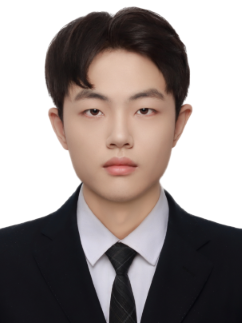}}]{Wenxuan Yu}
received the B.E. degree in mechanical design, manufacturing, and automation from Chongqing University in 2025. He is currently pursuing the M.E. degree in mechanical engineering with Chongqing University, Chongqing, China. His research interests include computer vision, Gaussian splatting and deep learning.
\end{IEEEbiography}
\biogap

\begin{IEEEbiography}[{\includegraphics[width=1in,height=1.25in,clip,keepaspectratio]{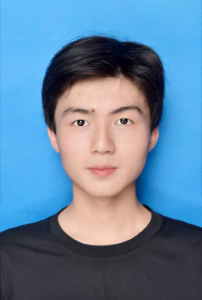}}]{Yicui Shi}
received the B.E. degree in automotive engineering from Chongqing University in 2025. He is currently pursuing the M.E. degree in automotive engineering with Chongqing University, Chongqing, China. His research interests include computer vision, Gaussian splatting and deep learning.
\end{IEEEbiography}
\biogap

\begin{IEEEbiography}[{\includegraphics[width=1in,height=1.25in,clip,keepaspectratio]{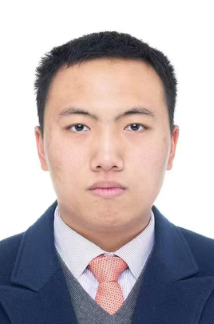}}]{Kunyang Huang}
is currently a master's degree student at Carnegie Mellon University. He previously interned at the Chongqing Institute of the Chinese Academy of Sciences and was a research intern at the Changan Automobile Research Institute. His research interests include autonomous driving perception and deep learning methods.
\end{IEEEbiography}
\biogap

\begin{IEEEbiography}[{\includegraphics[width=1in,height=1.25in,clip,keepaspectratio]{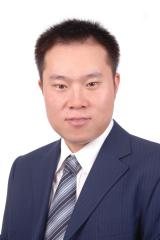}}]{Wenbo Chu}
received the B.S. degree in automotive engineering from Tsinghua University, China, in 2008, the M.S. degree in automotive engineering from RWTH Aachen University, Germany, and the Ph.D. degree in mechanical engineering from Tsinghua University, China, in 2014. He is currently a research fellow with Western China Science City Innovation Center of Intelligent and Connected Vehicles (Chongqing) Co., Ltd., and the National Innovation Center of Intelligent and Connected Vehicles.
\end{IEEEbiography}
\biogaplarge

\begin{IEEEbiography}[{\includegraphics[width=1in,height=1.25in,clip,keepaspectratio]{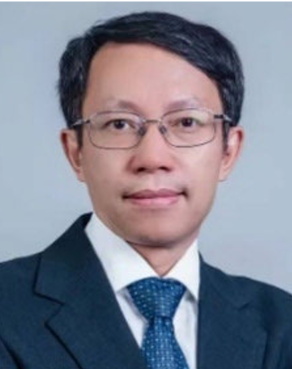}}]{Keqiang Li}
received the B.E. degree from Tsinghua University, Beijing, China, in 1985, and the M.E. and Ph.D. degrees from Chongqing University, Chongqing, China, in 1988 and 1995, respectively. He is currently a Professor with the School of Vehicle and Mobility, Tsinghua University. He is the Chief Scientist of Intelligent and Connected Vehicle Innovation Center of China, and the Director of the State Key Laboratory of Automotive Safety and Energy of China. His current research interests include intelligent connected vehicles, cloud-based control for vehicles, and vehicle dynamics systems.
\end{IEEEbiography}

\end{document}